\documentclass[twocolumn]{scholar7}

\usepackage{enumitem}

\usepackage{multirow}
\usepackage{placeins}
\usepackage{booktabs}
\usepackage{tabularx}
\usepackage{makecell}   
\usepackage{algorithm}
\usepackage{algorithmic}
\usepackage{xcolor}
\usepackage{amsthm}
\usepackage{rotating}
\newtheorem{proposition}{Proposition}[section]

\newcommand{\gray}[1]{\textcolor{gray}{#1}}
\newcommand{\best}[2]{\textbf{#1} $\pm$ \textbf{#2}}
\newcommand{\OTx}{\mathrm{OT}(x)}
\newcommand{\OTy}{\mathrm{OT}(y)}

\title{DISCOVER: A Solver for Distributional Counterfactual Explanations}

\author[1]{Yikai Gu}
\author[2]{Lele Cao}
\author[3]{Bo Zhao}
\author[4]{Lei Lei}
\author[1]{Lei You}

\affiliation[1]{Technical University of Denmark}
\affiliation[2]{Scholar7}
\affiliation[3]{Aalto University}
\affiliation[4]{Xi'an Jiaotong University}

\correspondence{\email{leiyo@dtu.dk}}
\date{June 2026}

\abstract{%
Counterfactual explanations (CE) explain model decisions by identifying input modifications that lead to different predictions. Most existing methods operate at the instance level. Distributional Counterfactual Explanations (DCE) extend this setting by optimizing an optimal transport objective that balances proximity to a factual input distribution and alignment to a target output distribution, with statistical certification via chance constrained bounds. However, DCE relies on gradient based optimization, while many real-world tabular pipelines are dominated by non-differentiable models. We propose DISCOVER, a model-agnostic solver for distributional counterfactual explanations. DISCOVER preserves the original DCE objective and certification while replacing gradient descent with a budgeted propose-and-select search paradigm. It exploits a sample-wise decomposition of the transport objective to compute per-row impact scores and enforce a top-$k$ intervention budget, focusing edits on the most influential samples. To guide candidate generation without predictor gradients, DISCOVER introduces an OT-guided cone sampling primitive driven by input-side transport geometry. Experiments on multiple tabular datasets demonstrate strong joint alignment of input and output distributions, extending distributional counterfactual reasoning to modern black box learning pipelines. A code repository is available at: \texttt{\small \url{https://github.com/VALHALLA9/Discover}}
}

\begin{document}

\maketitle

\section{Introduction}

\subsection{Background and Motivations}

Counterfactual explanations (CE) are widely used in explainable artificial intelligence (XAI) to provide actionable insights for automated decision systems \cite{dwivedi2023explainable,verma2024counterfactual}.
They describe minimal and feasible changes to the input that would lead a model to produce a desired outcome \cite{wachter2017counterfactual}.
Most CE methods operate at the instance level and are designed to provide recourse for a single instance, often incorporating feasibility constraints and user-specific costs \cite{ustun2019actionable,karimi2020model} or returning multiple diverse alternatives \cite{mothilal2020explaining}.
However, many real-world applications require population-level interventions rather than isolated edits.
For example, policy makers may ask how to shift the risk distribution of a subgroup, or financial institutions may seek systematic recourse strategies that remain consistent with the overall data manifold.
Existing global or group-level recourse toolkits partially address this issue by aggregating or organizing per-instance actions \cite{rawal2020beyond,ley2023globe}, but their objectives remain defined over individual edits rather than distributions, and they do not directly optimize the shape of the model’s output distribution that stakeholders aim to control. These settings call for distributional counterfactual explanations, where the goal is to generate a counterfactual distribution that both remains close to the factual input distribution and achieves a desired shift in model outcomes.

Distributional Counterfactual Explanations (DCE) provides a formulation of this problem using optimal transport \cite{you2025distributional}. DCE balances an input-side transport cost that preserves plausibility with an output-side transport cost that enforces target predictions, together with statistical certification via chance-constrained bounds. This distributional perspective offers a principled alternative to instance-level recourse by explicitly reasoning about population structure and group-level feasibility.

Despite this, the practical adoption of DCE solver remains limited by its optimization machinery. This gradient-based algorithm relies on differentiability assumptions and computation graphs, which restrict its applicability to smooth predictive models. While gradient-based optimization is well supported for differentiable models \cite{glorot2011deep,baydin2018automatic}, practical tabular systems are often non-differentiable or black-box systems, including tree ensembles, rule-based preprocessing, and mixed discrete-continuous feature constraints \cite{breiman2001random,friedman2001greedy,shwartz2022tabular,erickson2025tabarena}. In such settings, gradient based distributional optimization becomes unreliable or infeasible.

More importantly, simply replacing gradients with random sampling is not sufficient. Distributional counterfactual generation is a highly non-convex problem over mixed-type spaces, where naive derivative-free search can be prohibitively inefficient and unstable at the population scale. This motivates the need for a solver that is model-agnostic while still exploiting the structure of optimal transport objectives.

\subsection{Main Contributions}

In this work, we propose \textbf{DISCOVER}, a model-agnostic solver for distributional counterfactual explanations that preserves the certified DCE objective while rethinking the optimization paradigm. Our key insight is that optimal transport based distributional objectives admit a practical sample-level structure: global transport shifts can often be attributed to a small subset of influential samples. A comparison of existing counterfactual paradigms and their properties is summarized in Table~\ref{tab:cf_paradigm}.

DISCOVER operationalizes this insight through per-sample impact scoring and a top-$k$ budget that localizes and edits the most influential individuals, yielding an interpretable form of distribution-level intervention.

Building on this structure, DISCOVER formulates distributional counterfactual generation as an iterative propose-and-select search over candidate counterfactual distributions, where multiple candidate updates are generated and the candidate minimizing the certified DCE objective is selected.

DISCOVER further improves black-box search by introducing an input-side OT guidance primitive that exploits the geometry induced by the input-side transport discrepancy to steer proposals toward high-plausibility regions without querying predictor gradients.

\begin{itemize}\setlength\itemsep{0.25em}
\item \emph{Budgeted Distributional Editing via OT Decomposition}: We reformulate the distributional alignment problem by uncovering a sample-wise decomposition of the OT objective. This allows us to introduce a top-$k$ impact budget, localizing interventions to the most influential samples for updates.
\item \emph{Modular Propose-and-Select Architecture}: We propose a solver-agnostic framework that separates candidate generation from objective certification. This design enables certified distributional counterfactual generation in non-smooth and mixed-type spaces where traditional gradient-based methods fail.
\item \emph{Predictor-Gradient-Free Geometric Guidance}: We introduce an OT-guided cone sampling primitive that leverages input-side transport gradients. This mechanism provides directed, geometry-aware proposals for black-box models, significantly improving search efficiency over naive derivative-free methods.
\item \emph{Unified Solution for Black-Box Pipelines}: We demonstrate that DISCOVER achieves strong joint alignment of input and output distributions across differentiable as well as non-differentiable predictors, effectively extending certified distributional reasoning to modern tabular pipelines.
\end{itemize}

\begin{table*}[t]
\vspace{-1mm}
\centering
\small
\setlength{\tabcolsep}{3pt}
\renewcommand{\arraystretch}{1.15}
\caption{Comparison of counterfactual paradigms. DISCOVER keeps DCE's distribution-level objective and certification, while enabling model-agnostic optimization with explicit top-$k$ budgeted editing.}
\vspace{-2mm}
\begin{tabular}{lcccc}
\toprule
Method & Dist.\ goal & Certified & Model-Agnostic & Top-$k$ \\
\midrule
Instance CE & No & No & \makecell{Often\\Yes} & No \\
Group/Global CE & Partial & No & \makecell{No} & No \\
DCE~\cite{you2025distributional} & Yes & Yes & No & No \\
DISCOVER (ours) & Yes & Yes & Yes & Yes \\
\bottomrule
\end{tabular}
\label{tab:cf_paradigm}
\vspace{-3mm}
\end{table*}

\section{Distributional Counterfactual Formulation and Limitations}
Here we briefly review the distributional counterfactual formulation introduced by DCE to establish notation and certification semantics. We then describe the combined objective used to operationalize the chance constraints, before discussing why the original gradient-based optimization becomes problematic in realistic tabular pipelines. This sets the stage for the solver design.

\subsection{Problem Formulation}
\label{sec:objective}
Following DCE, we use $P$ to denote the scalar satisfaction probability of the chance constraints. This scalar is not a probability distribution. We use $\mathbb{P}[\cdot]$ to denote the probability of an event. 
Let $b:\mathcal{R}^d \to \mathcal{R}$ be a black-box model. Given a factual input distribution $\mathbf{x}'$ with outputs $y' = b(\mathbf{x}')$ and a target output distribution $y^\ast$, DCE seeks a counterfactual input distribution $\mathbf{x}$ that remains close to $\mathbf{x}'$ while aligning the model output distribution $b(\mathbf{x})$ with $y^\ast$. The problem is formulated over empirical samples and enforces robustness through chance constraints on both the input and output sides.

To measure input proximity, DCE employs the sliced Wasserstein distance $\mathcal{SW}^2$, which remains scalable in high dimensions, while output alignment is measured using the Wasserstein distance $\mathcal{W}^2$, capturing discrepancies at the distributional level rather than only in expectation \cite{you2025distributional}.

\begin{align}
\label{eq:dce}
\text{[DCE Problem]}\quad & \max_{\mathbf{x},\,P}\ P \\
\label{eq:dce:b}
\text{s.t.}\quad & P \le \mathbb{P}\!\left[\mathcal{SW}^{2}(\mathbf{x},\mathbf{x}')<U_x\right] \\
\label{eq:dce:c}
& P \le \mathbb{P}\!\left[\mathcal{W}^{2}\!\big(b(\mathbf{x}),y^\ast\big)<U_y\right] \\
\label{eq:dce:d}
& P \ge 1-\frac{\alpha}{2}\,.
\end{align}

Constraint \eqref{eq:dce:b} enforces proximity between the candidate and factual input distributions, while \eqref{eq:dce:c} imposes the analogous requirement on the model outputs relative to $y^\ast$. The confidence level $\alpha$ in \eqref{eq:dce:d} guarantees joint satisfaction of both constraints with probability at least $1-\alpha/2$. This chance-constrained optimal-transport formulation provides interpretable, distribution-level alignment while remaining scalable to high-dimensional inputs via slicing.

\subsection{Distributional Counterfactual Objective}

Building on the formulation above, let $Q_x$ and $Q_y$ denote the input-side and output-side transport costs induced by sliced Wasserstein and Wasserstein distances, respectively. To operationalize the chance-constrained problem, DCE combines these two costs through an objective with a trade-off parameter $\eta \in [0,1]$,

\begin{equation}
Q(x,\mu,\nu,\eta) \triangleq (1-\eta)\,Q_x(x,\mu) + \eta\,Q_y(x,\nu),
\end{equation}
where $\mu$ and $\nu$ denote the corresponding optimal transport plans. For suitable choices of $\eta$, minimizing $Q$ yields candidate solutions that satisfy the empirical chance constraints in \eqref{eq:dce:b}--\eqref{eq:dce:d} with high probability, highlighting the role of balancing input proximity and output alignment.

However, optimizing this objective with gradient-based solvers can be problematic in practical tabular settings. Many tabular prediction systems rely on non-differentiable components such as tree-based models or rule-based preprocessing pipelines, which prevent reliable gradient access. A detailed discussion of these limitations is provided in Appendix B.

\section{DISCOVER: OT-Structured Sparse Distributional Search}
\label{ch:framework}
DISCOVER is a solver for Distributional Counterfactual Explanations that preserves the DCE objective and certification layer, while introducing an OT-structured optimization paradigm for black-box and mixed-type tabular pipelines.
Instead of treating distribution editing as a global gradient descent over all samples, DISCOVER explicitly exploits the sample-wise structure induced by optimal transport.
It performs budgeted distribution editing via top-$k$ impact selection, and optimizes the certified objective through a propose-and-select search over candidate counterfactual distributions.
All candidate distributions are evaluated by the same distribution-level objective $Q$ with the same certification mechanism as DCE.
DISCOVER does not require model gradients.
\subsection{DISCOVER Framework}
\label{sec:new_dce_framework}

\begin{algorithm}[t]
\caption{DISCOVER: Budgeted Propose-and-Select Distributional Solver}
\label{alg:discover}
\begin{algorithmic}[1]
\REQUIRE Factual sample $X'$, predictor $b(\cdot)$, target outputs $Y^\ast$, bounds $(U_x,U_y)$, confidence level $\alpha$,
top-$k$ budget $k$, candidates per iteration $M$, max iterations $T$
\ENSURE Edited sample $\widehat{X}$ or $\varnothing$

\STATE Initialize $X \leftarrow X'$
\FOR{$t = 1$ to $T$}
    \STATE Evaluate $Q(X;\eta)$ and update $\eta$ using the same certification and interval-narrowing rule as DCE
    \STATE Compute per-row impact scores $\{q_i\}_{i=1}^n$
    \STATE Select editable row set $I \leftarrow$ indices of the top-$k$ scores
    \STATE Compute input-side OT guidance signal $g$ from $\mathcal{SW}^2(X,X')$
    \STATE Set baseline candidate $X^{(0)} \leftarrow X$ (no-op update)

    \FOR{$m = 1$ to $M$}
        \STATE $\textsc{Optimizer}$ (Sec.~\ref{sec:strategy_layer}) proposes $X^{(m)}$ by editing rows in $I$ using the shared OT-guided cone sampling primitive (Sec.~\ref{sec:gradient_guided}) guided by $-g$
    \ENDFOR

    \STATE $\widehat{X} \leftarrow \arg\min_{m \in \{0,1,\dots,M\}} Q\!\big(X^{(m)};\eta\big)$
    \STATE $X \leftarrow \widehat{X}$
\ENDFOR
\STATE Check whether $\widehat{X}$ satisfies the same DCE certification criteria
\IF{certification holds}
    \STATE \textbf{return} $\widehat{X}$
\ELSE
    \STATE \textbf{return} $\varnothing$
\ENDIF
\end{algorithmic}
\end{algorithm}

At each iteration, DISCOVER first evaluates the current distribution using the same sliced Wasserstein distance on inputs and one-dimensional Wasserstein distance on outputs as in DCE. These distances are computed from empirical optimal transport plans and certified using the same trimmed quantile bands and upper-confidence limits. The balance parameter $\eta$ is updated using the same interval-narrowing rule. Therefore, DISCOVER changes neither the distributional objective nor the certification semantics.

\begin{figure}[!t]
  \centering
  \includegraphics[width=\linewidth]{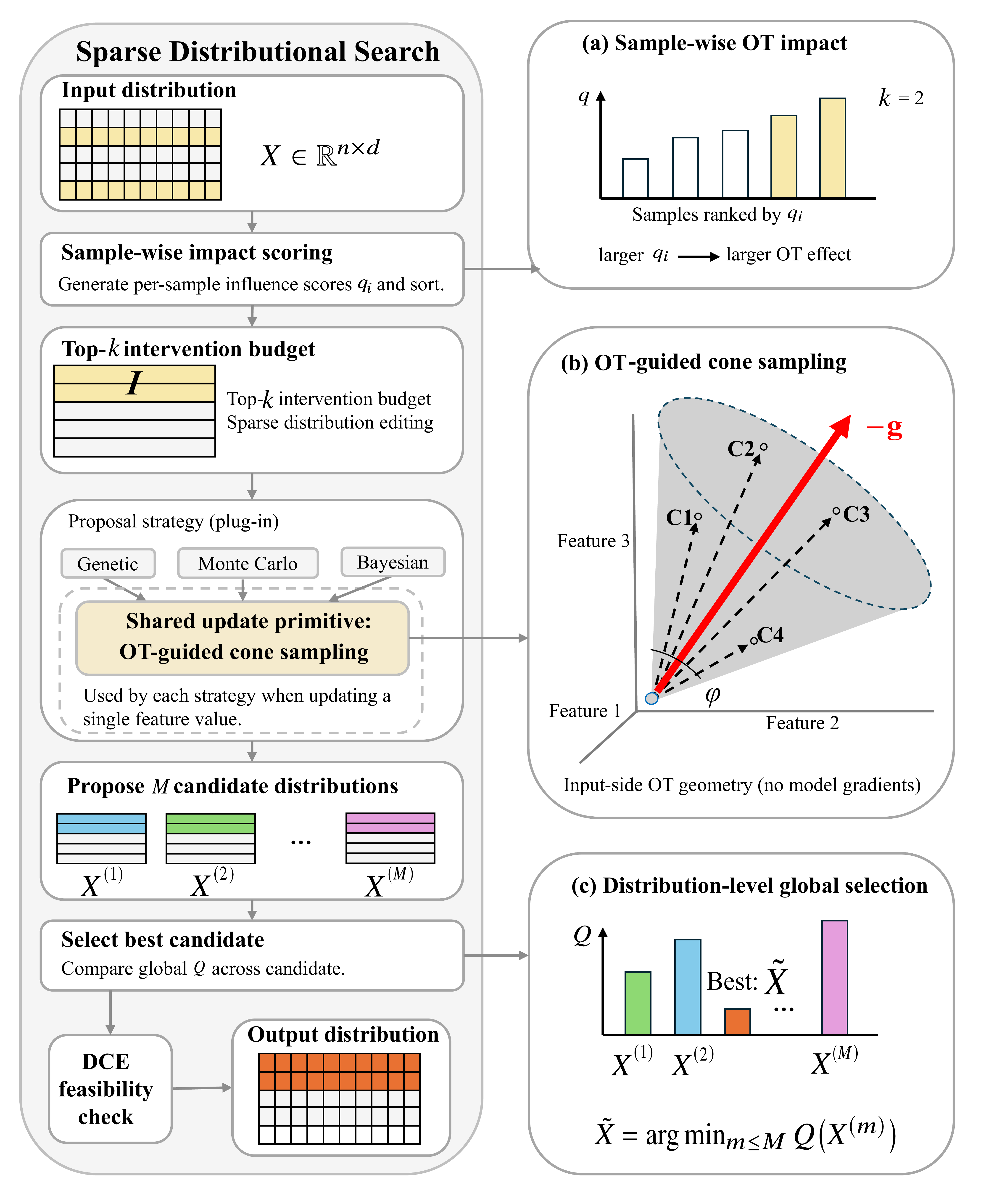}
      \caption{
      Overview of DISCOVER.
        DISCOVER preserves the DCE objective $Q$ and its certification layer.
        Each iteration computes per-sample impact scores $\{q_i\}$ from the current OT objective and selects a top-$k$ active set defining the intervention budget.
        It generates $M$ candidate counterfactual distributions by editing only these samples using a shared OT-guided cone sampling primitive.
        DISCOVER selects the candidate with the smallest certified objective $Q$ and repeats until the iteration budget is exhausted.
      }
    \label{fig:discover}
\end{figure}

The optimization challenge in tabular settings is not only the lack of model gradients, but also the need for budgeted interventions at the population level.
DISCOVER makes this budget explicit.
It attributes the distribution-level objective to sample-wise contributions and edits only a limited subset of influential instances in each iteration.

To operationalize this idea, DISCOVER decomposes the distributional objective into per-row impact scores.
On the input side, DCE induces OT plans over one-dimensional projections
$\Theta=\{\theta_k\}_{k=1}^{N}\subset\mathbb{S}^{d-1}$.
Under this plan, the input-side contribution of each sample $x_i$ can be written as
\begin{equation}
q_i^{(x)}
=
\frac{1}{N}\sum_{k=1}^{N}\sum_{j=1}^{n}
\bigl|\theta_k^\top x_i-\theta_k^\top x'_j\bigr|^{2}\,\mu^{(k)}_{ij},
\end{equation}
which measures how strongly $x_i$ contributes to the current input-side transport discrepancy.
An analogous decomposition holds on the output side under the transport plan $\nu$, yielding
\begin{equation}
q_i^{(y)}
=
\sum_{j=1}^{n}
\bigl|b(x_i)-y^\ast_j\bigr|^{2}\,\nu_{ij}.
\end{equation}
Combining both terms, the overall per-row impact score is defined as
\begin{equation}
q_i
=
(1-\eta)\,q_i^{(x)} + \eta\,q_i^{(y)}.
\end{equation}

\begin{proposition}[Row-wise decomposition of the certified objective]
\label{prop:row_decomp}
For fixed OT plans $\mu=\{\mu^{(k)}\}_{k=1}^N$ and $\nu$, the certified objective admits an exact row-wise decomposition
$
Q(x,\mu,\nu,\eta)=\sum_{i=1}^n q_i,
$
where $q_i^{(x)}$, $q_i^{(y)}$, and $q_i$ are defined above.
In particular, $Q_x(x,\mu)=\sum_{i=1}^n q_i^{(x)}$ and $Q_y(x,\nu)=\sum_{i=1}^n q_i^{(y)}$.
\end{proposition}
\noindent Proof is provided in Appendix D.

The combined score $q_i$ directly measures how strongly sample $x_i$ contributes to the current certified objective.
DISCOVER uses $q_i$ to define a top-$k$ editable set $I$ at each iteration.
Only rows in $I$ may change, and all other rows remain fixed.
This top-$k$ gate implements an explicit intervention budget.
It reduces the effective search space while keeping the objective unchanged.
The top-$k$ gate controls the support of each candidate update over empirical samples. The induced distributional change is not assumed to be sparse by construction. Instead, distribution-level behavior is controlled by the certified DCE objective: the input-side term $SW^2(\mathbf{x},\mathbf{x}')$ penalizes candidate distributions that move far from the factual distribution, while the output-side term $W^2(b(\mathbf{x}),y^\ast)$ encourages the predicted output distribution to match the target. Thus, a localized candidate edit is selected only when it improves the distribution-level transport objective and passes the DCE feasibility check.

Given the editable set $I$, DISCOVER performs a distribution-level propose-and-select search.
It generates $M$ candidate counterfactual distributions by editing only rows in $I$ with a shared proposal primitive.
Each candidate is evaluated by the same certified objective $Q(\cdot;\eta)$, and the best candidate is selected as the next iterate. We additionally include the current distribution as a baseline no-op candidate (Algorithm~\ref{alg:discover}), which guarantees that the certified objective does not increase within each iteration.

\begin{proposition}[Monotonicity of the propose-and-select step]
\label{prop:monotone_ps}
Let $X$ be the current iterate and let $$\widehat{X}=\arg\min_{m\in\{0,1,\dots,M\}} Q(X^{(m)};\eta)$$ be the selected candidate,
where $X^{(0)}=X$ is the no-op proposal and $\eta$ is fixed during the candidate evaluation.
Then $Q(\widehat{X};\eta)\le Q(X;\eta)$.
\end{proposition}
\noindent Proof is provided in Appendix D.

This explicit candidate selection is a key difference from single-path descent methods and is well-suited for non-convex and mixed-type optimization landscapes.

To bias proposals toward small input-side distance without querying predictive model gradients, DISCOVER computes an input-side OT guidance field from $\mathcal{SW}^2(X,X')$ and uses it only in the proposal step.
The output side remains fully black-box because no $\nabla b$ is required.
After the iteration budget, DISCOVER returns the best candidate that satisfies the same DCE certification criteria.

\subsection{A Modular Proposal Mechanism for Distributional Search}
\label{sec:strategy_layer}

DISCOVER separates the certified distributional objective from the mechanism that proposes candidate distributions.
The modular optimizer is a proposal layer.
It receives the current iterate $X$, the editable row set $I$ induced by the top-$k$ intervention budget, and a small set of exploration parameters.
It outputs $M$ candidate distributions $\{X^{(m)}\}_{m=1}^{M}$ that only edit rows in $I$ and satisfy domain constraints and actionability rules. All optimizers share a common proposal primitive, OT-guided cone sampling (Sec.~\ref{sec:gradient_guided}), and differ only in how they combine and apply it to generate candidates.

Different optimizers differ only in how they propose candidates.
All candidates are evaluated by the same certified objective $Q(\cdot;\eta)$, and DISCOVER selects the best candidate.
This makes the overall solver strategy-agnostic while keeping a single objective and a single certification protocol.

\textbf{Monte Carlo Optimizer.}
The Monte Carlo optimizer proposes $M$ validity-preserving candidate edits around the current iterate.
The top-$k$ gate confines changes to the most influential rows, and an input-side OT guidance field $g=\nabla_X \mathcal{SW}^{2}(X,X')$ biases proposal directions.
Numerical features are perturbed within valid ranges, and categorical features are modified within admissible levels.
Among the $M$ candidates, DISCOVER selects the one with the smallest certified objective value.

\textbf{Genetic Optimizer.}
The genetic optimizer produces candidates through recombination and guided mutation on the same editable set $I$.
Crossover reuses partial structures from the current iterate, while mutations introduce local variations in a domain--preserving manner.
Both operations are biased by the same input--side guidance field $g$.
Each candidate is evaluated by $Q(\cdot;\eta)$, and DISCOVER selects the best candidate.

\subsection{OT-Guided Cone Sampling Primitive}
\label{sec:gradient_guided}

Edits are restricted to the current top-$k$ rows $I$, so that each iteration focuses on the samples under the intervention budget.
To guide candidate generation without relying on model gradients, DISCOVER computes a single input-side guidance field
\[
-g \;=\; -\nabla_x \mathcal{SW}^{2}(x,x'),
\]
which depends only on the sliced Wasserstein distance between the current distribution and the factual reference.
This field captures geometry of input-side transport under the current OT structure.
It is independent of the predictive model, and $\nabla b$ is never queried.

The guidance field $g$ is used only to bias proposal directions.
Candidate updates are sampled within a cone of half-angle $\phi$ around the direction $-g$, take bounded steps, and are projected back to the valid domain.
No gradient descent is performed on the full objective $Q$, and $\nabla b$ is never queried.
The guidance field is evaluated once per outer iteration and reused across all $M$ candidate trials generated by the modular optimizer, ensuring low overhead while maintaining consistent directional bias.

\begin{algorithm}[h]
\caption{OT-guided cone sampling for numerical features}
\label{alg:cone-cont}
\begin{algorithmic}[1]
\STATE \textbf{Input:} row index $i$, current row $x_{i\cdot}$, editable indices $\mathcal{R}$,
guidance $g$, feature ranges $\{[R_p^{\min},R_p^{\max}]\}_{p\in\mathcal{R}}$
\STATE \textbf{Output:} updated row $x_{i\cdot}$ on $\mathcal{R}$
\FOR{$p \in \mathcal{R}$}
    \STATE sample direction  $d_{ip}$ in a cone around $-g$ (half-angle $\phi$)
    \STATE sample a step size $\lambda \in [0,\lambda_{\max}]$
    \STATE update
    $
    x_{ip} \gets \Pi_{[R_p^{\min},R_p^{\max}]}
    \Bigl(x_{ip} + \lambda (R_p^{\max}-R_p^{\min}) \, d_{ip}\Bigr)
    $
\ENDFOR
\STATE \textbf{return} updated $x_{i\cdot}$
\end{algorithmic}
\end{algorithm}

\textbf{Numerical coordinates.}
For numerical features, DISCOVER generates proposals using a shared OT-guided cone sampling primitive.
Each editable row is updated by sampling a direction that is biased toward the negative input-side guidance field $-g$,
where $g=\nabla_X \mathcal{SW}^2(X,X')$ depends on the transport discrepancy to the factual distribution.
This provides a lightweight geometric prior that favors minor edits in the input space, without requiring any gradient information from the predictive model.

The cone half-angle $\phi$ controls the concentration of proposals around this OT-induced direction, interpolating between focused transport-consistent moves and broader exploration.
A bounded step size $\lambda \in [0,\lambda_{\max}]$ is drawn and scaled by the feature range $(R_p^{\max}-R_p^{\min})$ to normalize updates across heterogeneous units.
Each proposed update is finally projected onto the valid interval $[R_p^{\min}, R_p^{\max}]$, ensuring feasibility by construction.
This procedure yields numerical candidate distributions that preserve input-side plausibility while enabling effective distribution-level search under the top-$k$ intervention budget.

\begin{algorithm}[h]
\caption{OT-guided cone sampling for categorical features}
\label{alg:cone-cat}
\begin{algorithmic}[1]
\STATE \textbf{Input:} row index $i$, current row $x_{i\cdot}$, editable indices $\mathcal{C}$,
guidance $g$, temperature $\tau$
\STATE \textbf{Output:} updated row $x_{i\cdot}$ on $\mathcal{C}$
\FOR{$p \in \mathcal{C}$}
    \STATE represent category $x_{ip}$ as a one-hot vector $o_{ip} \in \{0,1\}^{|\mathcal{V}_p|}$
    \STATE embed $z_{ip} \gets E_p[x_{ip}]$ where $E_p \in \mathbb{R}^{|\mathcal{V}_p|\times r}$ is a fixed embedding table (initialized once and kept constant)
    \STATE take a cone-biased embedding step:
    $\tilde z_{ip} \gets z_{ip} + \Delta$, where $\Delta \in \mathbb{R}^r$ is sampled in a cone around $-g$ (half-angle $\phi$)
    \STATE include the no-op option $v=x_{ip}$ in the admissible set under actionability constraints
    \STATE decode by sampling a category with
    $
    \Pr(v) \propto \exp\!\left(-\|E_p[v] - \tilde z_{ip}\|^2 / \tau\right)
    $
    under actionability constraints
\ENDFOR
\STATE \textbf{return} updated $x_{i\cdot}$
\end{algorithmic}
\end{algorithm}

\textbf{Categorical coordinates.}
For categorical features, DISCOVER performs proposals in a numerical embedding space where directions are meaningful.
For each categorical feature $p$, we construct a fixed embedding table $E_p \in \mathbb{R}^{|\mathcal{V}_p|\times r}$ once before optimization and keep it constant throughout.
In our implementation, $E_p$ is randomly initialized with a fixed random state for reproducibility, and $r$ is chosen by a deterministic rule based on the category cardinality $|\mathcal{V}_p|$ unless specified.
A cone-biased step guided by $-g$ is then applied in the embedding space.
To decode back to a discrete category, we compute distances between the updated embedding and all category embeddings and sample with a temperature-controlled rule.
Specifically, the probability of category $v$ is proportional to $\exp(-\|E_p[v] - \tilde z\|^2/\tau)$, where $E_p[v]\in\mathbb{R}^r$ denotes the embedding vector of category $v$.
The temperature $\tau$ controls the exploration and exploitation trade-off.
Finally, decoding is restricted to admissible categories under predefined domain and actionability constraints.

\section{Experimental Setup and Results}
\label{ch:Experiment}
\subsection{Experimental Setup}\label{sec:exp-setup}

We evaluate DISCOVER mainly on five tabular datasets: HELOC~\cite{FICO2018}, COMPAS~\cite{larson2016compas}, Hotel Booking~\cite{antonio2019hotel}, German Credit~\cite{hofmann1994germancredit}, and Cardiovascular Disease~\cite{halder2020cardio}. Quantitative comparisons are performed on HELOC, COMPAS and German Credit, where DISCOVER is compared with existing CE methods, including DCE~\cite{you2025distributional}, DiCE~\cite{mothilal2020explaining}, AReS~\cite{rawal2020beyond}, and GLOBE~\cite{ley2023globe}.
DiCE generates counterfactuals independently for individual instances,
whereas AReS and GLOBE operate at the group level by aggregating recourse patterns across samples. DCE is included as a gradient-based baseline in settings where differentiability is available,
enabling a direct comparison between gradient-based optimization and the proposed model-agnostic solver
under the same distributional objectives.

The main evaluation targets the distributional alignment objective inherited from DCE.
Let $\hat{\mathbf{x}}=\{\hat{x}_i\}_{i=1}^{n}$ denote the generated empirical counterfactual input distribution.
We report
\[
\mathrm{OT}(x)=\mathcal{SW}^{2}(\hat{\mathbf{x}},\mathbf{x}'), \qquad
\mathrm{OT}(y)=\mathcal{W}^{2}\!\left(b(\hat{\mathbf{x}}),y^\ast\right),
\]
where $\mathbf{x}'$ denotes the factual input distribution and $y^\ast$ denotes the target output distribution.
Thus, $\mathrm{OT}(x)$ measures input-side proximity to the factual distribution, while $\mathrm{OT}(y)$ measures output-side alignment to the target distribution.

We also report the maximum mean discrepancy between $\mathbf{x}'$ and $\hat{\mathbf{x}}$.
Using a kernel $\kappa$, its empirical squared form is
\begin{align*}
\mathrm{MMD}^{2}(\mathbf{x}',\hat{\mathbf{x}})
&= \frac{1}{n^{2}}\sum_{i=1}^{n}\sum_{j=1}^{n}\kappa(x'_i,x'_j) \\
&\quad + \frac{1}{n^{2}}\sum_{i=1}^{n}\sum_{j=1}^{n}\kappa(\hat{x}_i,\hat{x}_j) \\
&\quad - \frac{2}{n^{2}}\sum_{i=1}^{n}\sum_{j=1}^{n}\kappa(x'_i,\hat{x}_j).
\end{align*}
Finally, AReS Cost is computed using the cost function from AReS and is included only as a reference intervention-cost metric.
Lower values indicate smaller distances or lower intervention cost for the corresponding metric.

Ablations test the necessity of DISCOVER's solver components:
(i) sample-wise OT impact scoring and the top-$k$ intervention budget,
(ii) propose-and-select candidate search over multiple proposals per iteration,
and (iii) OT-guided cone sampling as a shared proposal primitive.
Extended ablations, runtime and scalability analysis, and additional comparisons with model-agnostic baselines are reported in Appendix. Details on the empirical distribution size, target construction, classification threshold, random seeds, and configuration files are provided in Appendix G.

\subsection{Main Results}\label{sec:main-results}

\begin{table*}[t]
\centering
\footnotesize
\setlength{\tabcolsep}{1.5pt}
\renewcommand{\arraystretch}{0.9}
\setlength{\abovecaptionskip}{2pt}
\caption{
Comparison of DISCOVER with counterfactual baselines on tabular datasets.
$\OTx$ and $\OTy$ denote input and output Wasserstein distances (lower is better).
Results report mean $\pm$ 80\% CI across SVM, MLP, RF, XGBoost and LightGBM.
AReS Cost is reported for reference only (not optimized by DISCOVER).
}
\label{tab:main-results}
\resizebox{\textwidth}{!}{
\begin{tabular}{llcccc}
\toprule
Dataset & Method & $\OTx$ & $\OTy$ & MMD & AReS Cost \\
\midrule
\multirow{5}{*}{\makecell{COMPAS \\ (Diff.)}}
& AReS     & \gray{$0.022 \pm 0.057$} & \gray{$0.539 \pm 0.043$} & \gray{$0.026 \pm 0.071$} & \gray{$2.076 \pm 2.038$} \\
& GLOBE    & \gray{$20.517 \pm 53.679$} & \gray{$0.009 \pm 0.027$} & \gray{$0.242 \pm 0.050$} & \gray{$1.070 \pm 1.808$} \\
& DiCE     & $0.112 \pm 0.086$ & $0.227 \pm 0.051$ & \best{0.055}{0.013} & \best{3.935}{0.006} \\
& DCE      & $0.093 \pm 0.022$ & $0.158 \pm 0.030$ & $0.089 \pm 0.021$ & $4.068 \pm 1.167$ \\
& DISCOVER & \best{0.068}{0.025} & \best{0.147}{0.058} & $0.068 \pm 0.020$ & $4.037 \pm 0.636$ \\
\midrule

\multirow{5}{*}{\makecell{COMPAS \\ (Non-diff.)}}
& AReS     & \gray{$0.017 \pm 0.020$} & \gray{$0.499 \pm 0.135$} & \gray{$0.028 \pm 0.035$} & \gray{$0.928 \pm 0.892$} \\
& GLOBE    & $0.354 \pm 0.028$ & $0.248 \pm 0.017$ & $0.203 \pm 0.026$ & \best{1.425}{0.852} \\
& DiCE     & $0.067 \pm 0.005$ & $0.238 \pm 0.033$ & $0.087 \pm 0.006$ & $ 2.281 \pm 0.452$ \\
& DCE      & -- & -- & -- & -- \\
& DISCOVER & \best{0.059}{0.002} & \best{0.141}{0.047} & \best{0.084}{0.010} & $2.922 \pm 0.098$ \\
\midrule

\multirow{5}{*}{\makecell{HELOC \\ (Diff.)}}
& AReS     & \gray{$0.001 \pm 0.001$} & \gray{$0.296 \pm 0.005$} & \gray{$0.000 \pm 0.000$} & \gray{$1.775 \pm 4.857$} \\
& GLOBE    & \gray{$1.600 \pm 0.290$} & \gray{$0.117 \pm 0.030$} & \gray{$0.006 \pm 0.000$} & \gray{$2.581 \pm 0.067$} \\
& DiCE     & $0.022 \pm 0.018$ & $0.164 \pm 0.061$ & $0.005 \pm 0.000$ & \best{5.534}{5.129} \\
& DCE      & $0.120 \pm 0.141$ & \best{0.138}{0.107} & $0.015 \pm 0.007$ & $6.323 \pm 0.232$ \\
& DISCOVER & \best{0.014}{0.011} & $0.190 \pm 0.025$ & \best{0.003}{0.002} & $9.485 \pm 0.194$ \\
\midrule

\multirow{5}{*}{\makecell{HELOC \\ (Non-diff.)}}
& AReS     & \gray{$0.001 \pm 0.000$} & \gray{$0.518 \pm 0.197$} & \gray{$0.001 \pm 0.000$} & \gray{$0.074 \pm 0.004$} \\
& GLOBE    & \gray{$6.663 \pm 5.504$} & \gray{$0.114 \pm 0.082$} & \gray{$0.060 \pm 0.051$} & \gray{$2.581 \pm 0.125$} \\
& DiCE     & $0.070 \pm 0.041$ & $0.232 \pm 0.042$ & $0.046 \pm 0.037$ & \best{3.932}{2.039} \\
& DCE      & -- & -- & -- & -- \\
& DISCOVER & \best{0.051}{0.048} & \best{0.160}{0.035} & \best{0.032}{0.026} & $3.962 \pm 1.933$ \\
\midrule

\multirow{5}{*}{\makecell{German Credit \\ (Diff.)}}
& AReS     & \gray{$0.018 \pm 0.055$} & \gray{$0.783 \pm 1.247$} & \gray{$0.031 \pm 0.089$} & \gray{$0.854 \pm 0.451$} \\
& GLOBE    & \gray{$230.847 \pm 468.610$} & \gray{$0.008 \pm 0.025$} & \gray{$0.250 \pm 0.547$} & \gray{$0.216 \pm 0.072$} \\
& DiCE     & $0.055 \pm 0.144$ & $ 0.110 \pm 0.288$ & $0.120 \pm 0.236$ & $4.563 \pm 10.168$ \\
& DCE      & \best{0.021}{0.023} & $0.521 \pm 0.439$ & \best{0.032}{0.013} & $5.867 \pm 17.498$ \\
& DISCOVER & $0.047 \pm 0.125$ & \best{0.103}{0.092} & $0.097 \pm 0.199$ & \best{2.858}{4.151} \\
\midrule

\multirow{5}{*}{\makecell{German Credit \\ (Non-diff.)}}
& AReS     & \gray{$0.003 \pm 0.003$} & \gray{$0.668 \pm 0.229$} & \gray{$0.008 \pm 0.007$} & \gray{$1.127 \pm 1.145$} \\
& GLOBE    & $0.030 \pm 0.005$ & $0.361 \pm 0.004$ & $0.052 \pm 0.014$ & \best{0.915}{0.144} \\
& DiCE     & $0.012 \pm 0.008$ & $0.364 \pm 0.183$ & \best{0.047}{0.030} & $5.058 \pm 0.717$ \\
& DCE      & -- & -- & -- & -- \\
& DISCOVER & \best{0.012}{0.007} & \best{0.229}{0.053} & $0.047 \pm 0.032$ & $6.946 \pm 1.738$ \\

\bottomrule
\end{tabular}
}
\end{table*}

\begin{figure}[t]
\centering
\begin{minipage}[t]{0.45\linewidth}
  \centering
  \includegraphics[width=\linewidth]{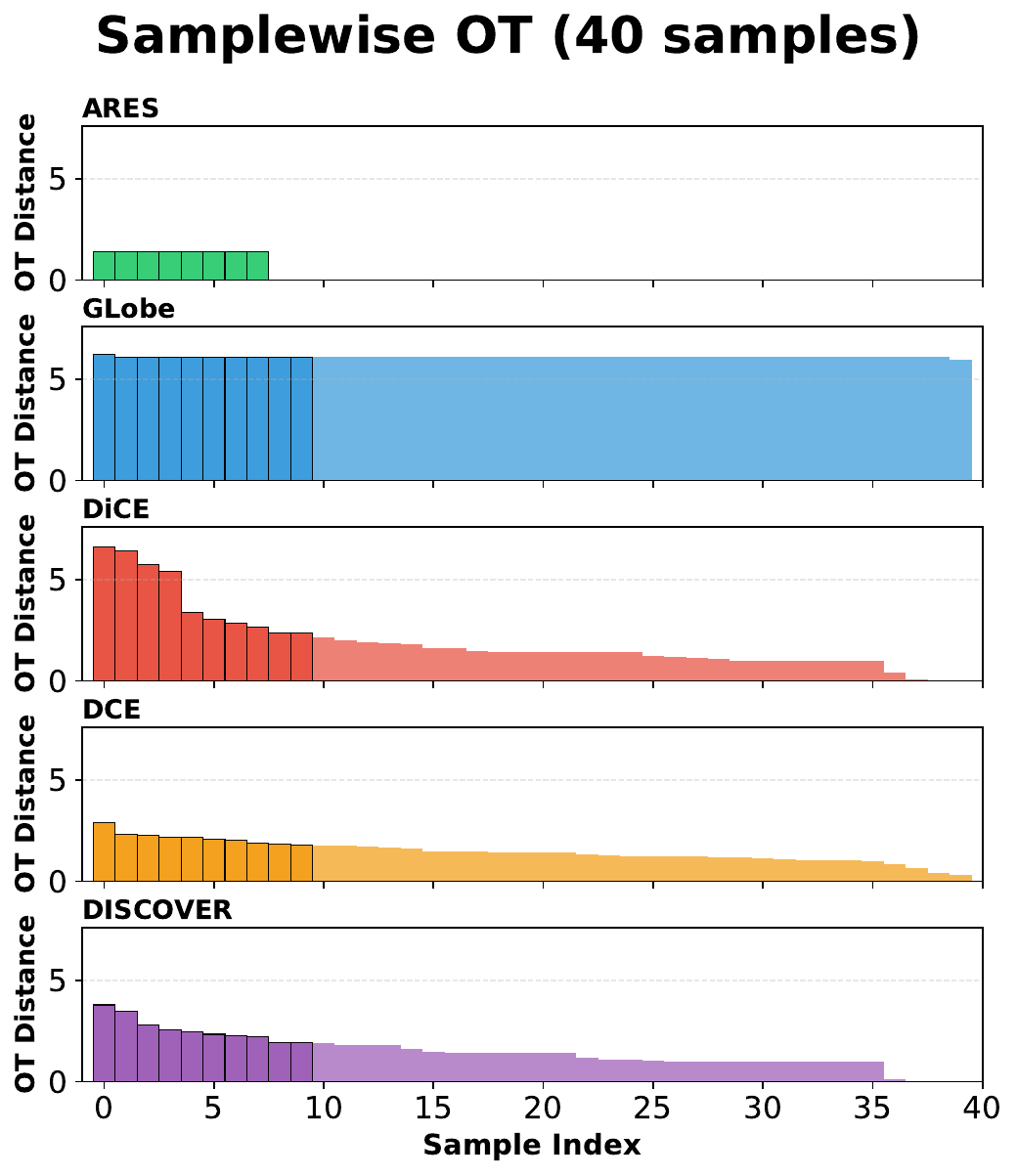}
  \caption{
    Per-sample input-side OT distances on COMPAS with an MLP.
    Distances are sorted in descending order within each method.
  }
  \label{fig:ot_comparison_compas_mlp}
\end{minipage}
\hfill
\begin{minipage}[t]{0.5\linewidth}
  \centering
  \includegraphics[width=\linewidth]{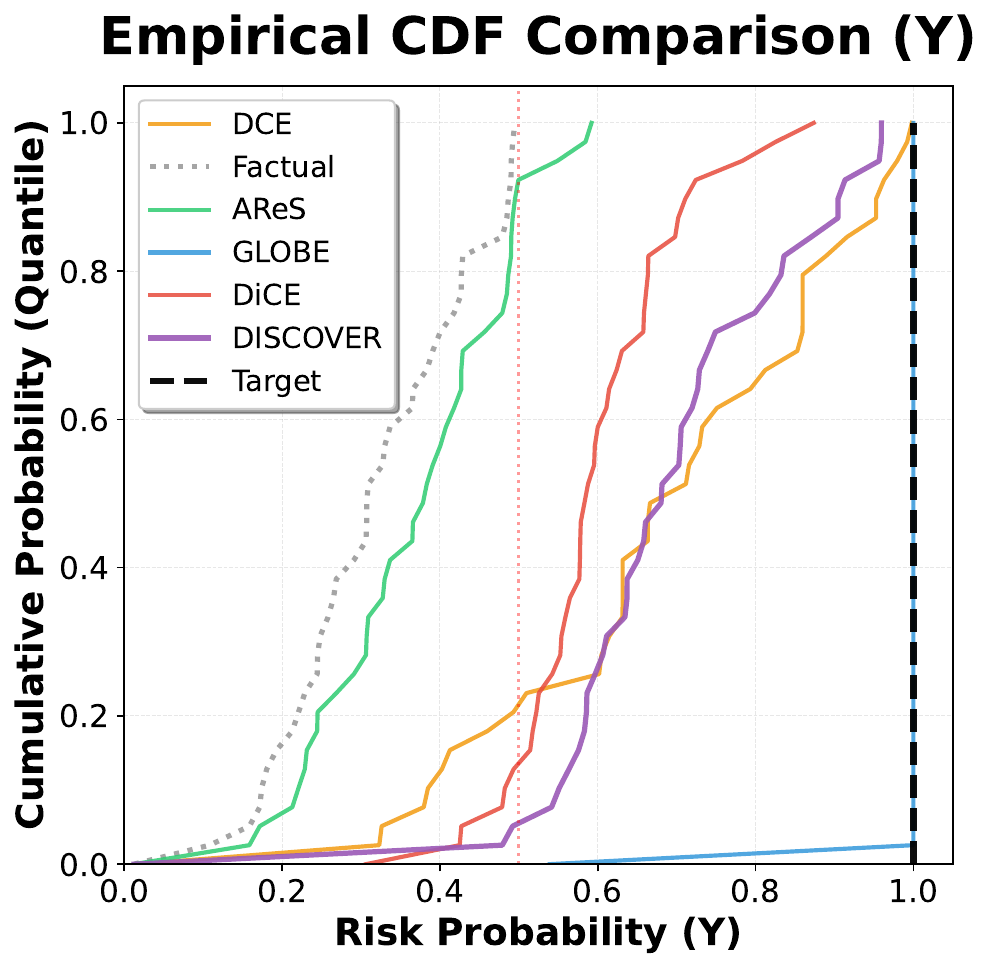}
  \caption{
    Empirical CDFs of model outputs on COMPAS with an MLP.
    Shown are factual outputs, the target distribution, and counterfactual outputs.
  }
  \label{fig:cdf_y_compas_mlp}
\end{minipage}
\end{figure}

Figures~\ref{fig:ot_comparison_compas_mlp} and~\ref{fig:cdf_y_compas_mlp} provide a qualitative comparison of input and output shift across methods.
Conservative methods can preserve the input distribution while failing to induce a meaningful output shift.
AReS produces uniformly small per-sample input-side transport distances, but its output CDF remains close to the factual distribution, leading to large $\OTy$.

In contrast, aggressive group-level translations can match the target outputs while producing large input-side transport deviations.
GLOBE aligns the output distribution closely to the target, but does so with substantially larger per-sample $\OTx$, indicating global distributional deviation.

DISCOVER achieves a sample-budgeted and coordinated intervention pattern.
It concentrates edits on a small subset of influential samples, yielding a controlled $\OTx$ profile while producing an output CDF that closely tracks the target.

We report the main quantitative comparison in Table~\ref{tab:main-results}.
Results are averaged over five predictive models (SVM, MLP, Random Forest, XGBoost, and LightGBM) and are reported separately for differentiable and non-differentiable settings.

We focus primarily on $\OTx$ and $\OTy$, which respectively measure the Wasserstein distance between the factual and counterfactual input and output distributions. These metrics directly reflect the quality of distributional counterfactual explanations, as they assess whether counterfactual edits simultaneously preserve the structure of the input distribution while steering model predictions toward the desired target distribution.

Across datasets, DISCOVER attains the best or second-best $\OTx$ in most settings and achieves the lowest $\OTy$ in all non-differentiable settings.

In contrast, several baseline methods exhibit imbalanced behavior between $\OTx$ and $\OTy$.
For instance, AReS frequently attains extremely small $\OTx$ and MMD values, suggesting minimal perturbations to the input distribution. However, this behavior is accompanied by substantially larger $\OTy$ values in nearly all cases, showing that the output distribution remains far from the target. This reflects the rule-based nature of AReS, which prioritizes conservative feature edits and feasibility constraints, often at the expense of achieving meaningful distributional shifts in model predictions. This leads to counterfactual distributions that remain close to the factual outputs under the OT criterion.

Similarly, DiCE, as an instance-based method, struggles to coordinate counterfactual edits across samples. While it may achieve reasonable $\OTx$ values in some settings, its $\OTy$ values are consistently higher than those of DISCOVER, indicating limited ability to align output distributions at the group level. This is consistent with the limitation of instance-wise generation when the evaluation target is defined over distributions.

\begin{table*}[t]
\centering
\footnotesize
\setlength{\tabcolsep}{1pt}
\renewcommand{\arraystretch}{0.95}
\caption{
Ablation of proposal guidance and editable feature types.
Results are averaged over models (treated as independent runs) and reported as mean $\pm$ 95\% confidence interval.
For each dataset, the best (lowest) value for each metric is highlighted in bold.
}
\label{tab:subset_ablation_single}

\resizebox{\textwidth}{!}{
\begin{tabular}{cc|cccc}
\toprule
Dataset & Method & $\OTx$ & $\OTy$ & MMD & AReS Cost\\
\midrule

\multirow{4}{*}{COMPAS}
& DISCOVER
& \textbf{0.1128 $\pm$ 0.0387}
& 0.0281 $\pm$ 0.0071
& \textbf{0.0461 $\pm$ 0.0120}
& \textbf{8.6432 $\pm$ 0.8135} \\
& Categorical only
& 0.1423 $\pm$ 0.0413
& 0.0297 $\pm$ 0.0060
& 0.0494 $\pm$ 0.0101
& 8.8009 $\pm$ 0.8645 \\
& numerical only
& 0.1183 $\pm$ 0.0390
& \textbf{0.0249 $\pm$ 0.0061}
& 0.0469 $\pm$ 0.0101
& 8.7429 $\pm$ 0.6693 \\
& No guidance
& 0.1409 $\pm$ 0.0406
& 0.0285 $\pm$ 0.0055
& 0.0497 $\pm$ 0.0081
& 9.1233 $\pm$ 0.6132 \\
\midrule

\multirow{4}{*}{Cardio}
& DISCOVER
& \textbf{0.0529 $\pm$ 0.0087}
& 0.0194 $\pm$ 0.0029
& \textbf{0.0223 $\pm$ 0.0015}
& \textbf{7.2161 $\pm$ 0.4619} \\
& Categorical only
& 0.0768 $\pm$ 0.0190
& \textbf{0.0140 $\pm$ 0.0040}
& 0.0268 $\pm$ 0.0017
& 7.4479 $\pm$ 0.4979 \\
& numerical only
& 0.0650 $\pm$ 0.0089
& 0.0206 $\pm$ 0.0031
& 0.0280 $\pm$ 0.0016
& 8.5790 $\pm$ 0.4547 \\
& No guidance
& 0.0909 $\pm$ 0.0180
& 0.0155 $\pm$ 0.0039
& 0.0326 $\pm$ 0.0021
& 8.7970 $\pm$ 0.5435 \\
\midrule

\multirow{4}{*}{HELOC}
& DISCOVER
& \textbf{0.0960 $\pm$ 0.0158}
& 0.0597 $\pm$ 0.0175
& \textbf{0.0280 $\pm$ 0.0018}
& \textbf{7.7486 $\pm$ 0.6226} \\
& Categorical only
& 0.1490 $\pm$ 0.0225
& 0.0600 $\pm$ 0.0188
& 0.0295 $\pm$ 0.0019
& 8.8170 $\pm$ 0.8046 \\
& numerical only
& 0.1097 $\pm$ 0.0183
& \textbf{0.0566 $\pm$ 0.0178}
& 0.0292 $\pm$ 0.0017
& 8.1970 $\pm$ 1.0586 \\
& No guidance
& 0.1605 $\pm$ 0.0252
& 0.0573 $\pm$ 0.0188
& 0.0295 $\pm$ 0.0016
& 9.6490 $\pm$ 0.8179 \\
\midrule

\multirow{4}{*}{German Credit}
& DISCOVER
& 0.1126 $\pm$ 0.0157
& 0.0430 $\pm$ 0.0246
& \textbf{0.0339 $\pm$ 0.0021}
& 7.5343 $\pm$ 0.2731 \\
& Categorical only
& 0.1893 $\pm$ 0.0336
& 0.0489 $\pm$ 0.0265
& 0.0357 $\pm$ 0.0017
& \textbf{7.2821 $\pm$ 0.2376} \\
& numerical only
& \textbf{0.1037 $\pm$ 0.0160}
& \textbf{0.0408 $\pm$ 0.0247}
& 0.0353 $\pm$ 0.0019
& 8.5443 $\pm$ 0.3417 \\
& No guidance
& 0.1631 $\pm$ 0.0249
& 0.0419 $\pm$ 0.0250
& 0.0377 $\pm$ 0.0018
& 8.3696 $\pm$ 0.2676 \\
\midrule

\multirow{4}{*}{Hotel Booking}
& DISCOVER
& \textbf{0.1081 $\pm$ 0.0226}
& 0.0257 $\pm$ 0.0092
& \textbf{0.0259 $\pm$ 0.0013}
& 16.4801 $\pm$ 2.7424 \\
& Categorical only
& 0.1936 $\pm$ 0.0732
& 0.0348 $\pm$ 0.0169
& 0.0260 $\pm$ 0.0013
& 18.2824 $\pm$ 3.3721 \\
& numerical only
& 0.1250 $\pm$ 0.0276
& \textbf{0.0192 $\pm$ 0.0077}
& 0.0270 $\pm$ 0.0013
& \textbf{15.6462 $\pm$ 2.5465} \\
& No guidance
& 0.1772 $\pm$ 0.0508
& 0.0291 $\pm$ 0.0155
& 0.0271 $\pm$ 0.0014
& 18.6048 $\pm$ 3.1646 \\
\bottomrule
\end{tabular}
}
\end{table*}

GLOBE exhibits high variance and occasionally extremely large $\OTx$ values (e.g., German Credit), while achieving very small $\OTy$.
This indicates that GLOBE can reach output alignment by applying large, near-uniform translations across many samples, which moves the empirical input distribution far from the factual manifold.
Because $\OTx$ is a distribution-level transport distance, a small number of failed or overly aggressive runs can dominate the mean and inflate the confidence interval.
We therefore interpret these results as evidence that translation-based group recourse can be unstable under mixed constraints, and we complement mean statistics with per-sample profiles in Figure~\ref{fig:ot_comparison_compas_mlp}.

The original DCE method is included as a baseline in differentiable settings.
Although DCE and DISCOVER optimize the same certified distributional objective, DISCOVER remains competitive with DCE and can sometimes achieve lower $\OTx$/$\OTy$ under the same evaluation budget.
On German Credit in the differentiable setting, DCE achieves a lower $\mathrm{OT}(x)$, whereas DISCOVER obtains a much lower $\mathrm{OT}(y)$, reflecting a different input-output trade-off under the same certified objective.
We attribute this behavior to the solver paradigm rather than a change of objective.
DCE follows a single-path gradient descent that updates all samples simultaneously, which can be sensitive to non-convexity, slicing approximations, and mixed-type constraints.
In contrast, DISCOVER performs budgeted edits and applies a propose-and-select search over multiple candidate distributions, providing a more robust update mechanism under stochastic and discrete proposal dynamics.
Consistent with this explanation, Figure~\ref{fig:OT} shows that OT-guided proposals stabilize optimization trajectories, while overly aggressive population updates (larger $k$) lead to unstable behavior.
Additional ablations on key solver components are reported in the  Appendix E.

\subsection{Ablation Study}\label{sec:ablation-strategy}

\begin{figure*}[t]
  \centering
  \includegraphics[width=\textwidth]{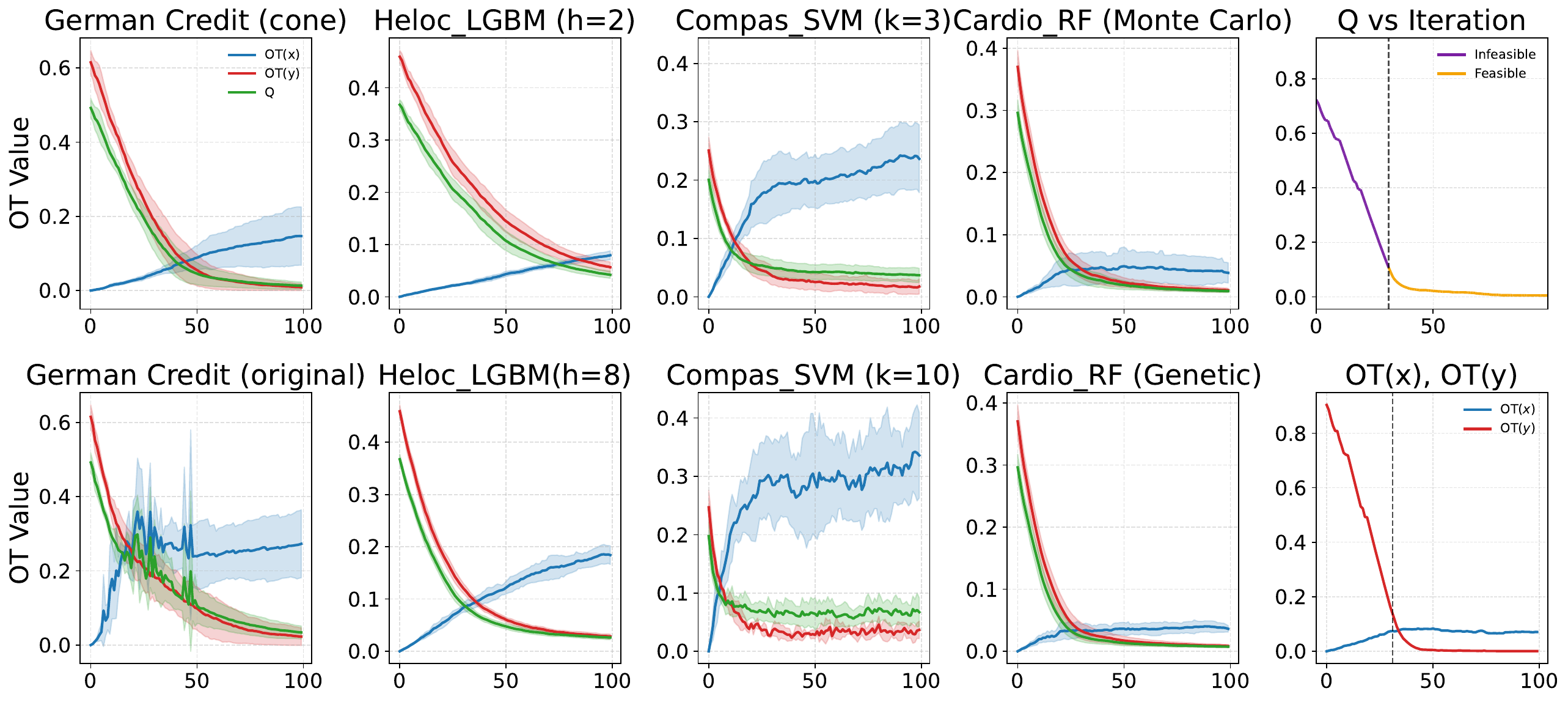}
    \caption{\normalfont
    Optimization dynamics of DISCOVER across datasets, models, and solver settings. 
    Each subplot reports input-side distance $\mathrm{OT}(x)$(blue), output-side distance $\mathrm{OT}(y)$(red), and the objective $Q$ over iterations. 
    Column 1 compares German Credit with OT-guided cone sampling and without cone guidance. 
    Column 2 compares different within-row update budgets $h$ on HELOC with LightGBM. 
    Column 3 compares different top-$k$ budgets on COMPAS with SVM. 
    Column 4 compares Monte Carlo and Genetic proposal strategies on Cardio with Random Forest. 
    Column 5 visualizes the transition from infeasible to feasible iterates. 
    Solid lines denote the mean over five random seeds, and shaded regions denote standard deviation.
    }
  \label{fig:OT}
\end{figure*}

We ablate solver components (Figure~\ref{fig:OT}) and editable feature subsets (Table~\ref{tab:subset_ablation_single}), and sweep $h$ and $k$ for stability--efficiency trade-offs under fixed budget.

Table~\ref{tab:subset_ablation_single} shows that allowing both numerical and categorical edits generally yields the lowest $\OTx$.
Restricting edits to a single type often increases $\OTx$, indicating reduced flexibility for transport-consistent updates.

For $\OTy$, numerical-only editing is often competitive (e.g., COMPAS, HELOC, German Credit, and Hotel Booking), while categorical edits matter on datasets where discrete switches are needed (e.g., Cardio).
Overall, enabling both types provides the most reliable balance between $\OTx$ and $\OTy$.
This is consistent with the objective structure: numerical edits follow the continuous geometry of $\mathcal{SW}^2$, whereas categorical edits introduce discrete moves that can be harder to optimize but sometimes necessary.

Numerical updates align well with the continuous geometry induced by $\mathcal{SW}^2$, while categorical updates introduce discrete jumps that can be harder to optimize but may be necessary to achieve target alignment under constraints.

Figure~\ref{fig:OT} summarizes optimization trajectories of $\OTx$, $\OTy$, and $Q$ across datasets, models, and solver settings.

Column 1 compares German Credit under two settings:
with OT-guided cone sampling (top) and without it (bottom).
When cone guidance is enabled, all three quantities decrease smoothly and remain stable.
Without guidance, $\mathrm{OT}(x)$ shows large fluctuations, which propagate to the objective $Q$.
As a result, both the final  $\OTx$ and $\OTy$ values and the objective are worse than in the guided case,
highlighting the stabilizing role of input-side geometric guidance.

The second column analyzes the effect of $h$, the number of features updated per selected sample,
on HELOC with LightGBM.
Increasing $h$ from $2$ to $8$ leads to faster convergence.
The variance of $Q$ and $\mathrm{OT}(y)$ is substantially larger for $h=2$,
while $h=8$ yields smoother trajectories.
Larger $h$ accelerates the decrease of $Q$ and $\OTy$, at the cost of a modest increase in $\OTx$ variance.

The third column studies the impact of $k$, the number of samples updated per iteration,
on COMPAS with an SVM.
With $k=3$, $\mathrm{OT}(x)$, $\mathrm{OT}(y)$, and $Q$ converge smoothly and remain stable.
In contrast, $k=10$ leads to large fluctuations and wide variance bands,
and the final solution is inferior across all metrics.
This supports the top-$k$ design: sample-level interventions stabilize the OT objective, while large-$k$ updates amplify coupling noise. We further examine whether the selected top-$k$ set is fixed by initialization. Appendix F.3 shows that top-$k$ membership changes over iterations, while the magnitude of the per-row scores generally decreases as optimization progresses.

The fourth column compares Monte Carlo and Genetic optimizers on Cardio with a Random Forest.
Both optimizers reach similar mean trajectories and final values of $Q$.
However, Monte Carlo exhibits noticeably larger variance,
reflecting higher stochasticity across runs.

The last column visualizes feasibility during optimization.
The top panel plots $Q$, where purple indicates infeasible iterates and yellow indicates feasible ones.
After finite iterations, marked by the vertical dashed line,
the optimization enters the feasible region.
The bottom panel shows that this transition coincides with a balance between $\mathrm{OT}(x)$ and $\mathrm{OT}(y)$,
confirming that feasibility is achieved when input proximity and output alignment are jointly satisfied.

\section{Conclusion}
In this paper, we introduced DISCOVER, a model-agnostic solver for distributional counterfactual explanations that preserves the certified optimal-transport objective of DCE. DISCOVER identifies influential samples through a sample-wise OT decomposition and applies a top-$k$ intervention budget to localize distributional edits. Across multiple tabular datasets and predictors, DISCOVER achieves strong alignment between input and output distributions. Crucially, these improvements are obtained without relying on predictive model gradients. As a modular and architecture-agnostic approach, DISCOVER integrates naturally with black-box tabular pipelines and broadens the applicability of certified distributional explanations.
\clearpage

\bibliographystyle{plainnat}
\bibliography{references}

\clearpage
\appendix

\section{Related Work}
\label{sec:ce}

Counterfactual explanations (CE) aim to provide actionable insight by identifying feasible input changes that alter a model prediction.
Wachter et al.~\cite{wachter2017counterfactual} formulate CE as an optimization problem that balances proximity to the factual instance and alignment with a target outcome, which has motivated a broad literature~\cite{guidotti2022counterfactual}.

\paragraph{Group-level, global, and transport-based recourse.}
Recent work also studies cohort-level and group counterfactual explanations. AReS~\cite{rawal2020beyond} and GLOBE~\cite{ley2023globe} summarize recourse patterns across groups, while Warren et al.~\cite{warren2024explaining} and Artelt and Gregoriades~\cite{artelt2024two} study multi-instance or group-level counterfactual explanations with an emphasis on coordinated recourse for multiple instances. Other work connects counterfactual reasoning with optimal transport: De Lara et al.~\cite{delara2024transport} study transport-based counterfactual models, Carrizosa et al.~\cite{carrizosa2024mathematical} formulate group counterfactual explanations through mathematical optimization for mixed tabular data, and Valero-Leal et al.~\cite{valeroleal2026optimal} study optimal transport maps and plans for groups and parameterized distributions, including geometry-preserving constraints beyond Wasserstein distance.

\paragraph{Model-agnostic and black-box counterfactual generation.}
A large body of work studies counterfactual generation for black-box predictors through model-agnostic objectives and derivative-free search.
Representative examples include MACE~\cite{yang2022mace}, NICE~\cite{brughmans2021nice}, which generates nearest-instance counterfactual explanations, and the branch-and-bound model-agnostic algorithm of Raimundo et al.~\cite{raimundo2024mining}, which mines Pareto-optimal counterfactual antecedents.
Other lines of work include reinforcement-learning-based approaches for scalable black-box recourse~\cite{samoilescu2021model}, causally constrained generation methods such as MC3G~\cite{dasgupta2025mc3g}, and plausibility-oriented tabular frameworks such as CountARFactuals~\cite{dandl2024countarfactuals}.
Genetic and sampling-based optimizers have also been used in other domains, such as molecule counterfactual generation~\cite{wellawatte2022model}.
These methods focus on instance-level recourse or antecedent search, and they do not optimize transport-based distances between empirical input and output distributions.

\paragraph{Distributional counterfactual explanations.}
Distributional Counterfactual Explanations (DCE)~\cite{you2025distributional} adopt a distribution-level formulation.
DCE seeks a counterfactual input distribution that steers the model output distribution toward a target while remaining close to the original population under optimal transport constraints~\cite{peyre2019computational}. It provides a certification mechanism for finite-sample feasibility.
The original DCE solver relies on gradient-based optimization and assumes differentiable predictors and pipelines, which restricts direct applicability to pipelines where the predictor and preprocessing are non-differentiable ~\cite{breiman2001random,friedman2001greedy,shwartz2022tabular,erickson2025tabarena}.

\paragraph{Distribution-aware counterfactual attribution.}
Related work has also leveraged distributional information to guide counterfactual explanations at the feature attribution level.
For example, \cite{lei2026joint} incorporate joint distribution--informed Shapley values to identify sparse and interpretable counterfactual attributions.
While sharing a distribution-aware perspective, these approaches focus on explaining and ranking feature-level interventions, rather than generating counterfactual input distributions or addressing solver design.

\paragraph{Positioning of DISCOVER.}
DISCOVER builds on the certified OT formulation of DCE while introducing a sparse, propose-and-select distributional search paradigm that does not query model gradients.
It makes the sample-level structure induced by transport objectives explicit through per-sample impact scoring, and supports black-box predictors through modular candidate proposals guided by input-side OT geometry.
This design extends certified distributional counterfactual explanations to realistic tabular pipelines while keeping the same OT objective and applying the same UCL-based feasibility check as in DCE.

\section{Limitations of Gradient-Based Optimization}

Tabular prediction tasks often rely on non-differentiable or only piecewise smooth models
\cite{shwartz2022tabular,erickson2025tabarena}.
Tree- and rule-based methods remain strong baselines in this setting, particularly for datasets
with mixed feature types, missing values, and structured constraints
\cite{breiman2001random,friedman2001greedy}.
Moreover, real-world tabular pipelines frequently include preprocessing steps such as
discretization, thresholding, and rule-based logic, which can render an otherwise
differentiable model non-differentiable as a whole.
As a result, many deployed tabular systems do not provide reliable input gradients.

However, the limitations of gradient-based solvers in distributional counterfactual generation
extend beyond the mere absence of gradients.
The original DCE solver performs joint updates of an empirical counterfactual distribution
through smooth descent steps, implicitly assuming that the optimization landscape is well
behaved under global, population-wide perturbations.
In realistic tabular settings, the distributional objective is highly non-convex and defined
over mixed continuous and categorical domains, where gradients can be unstable, noisy, or
misleading even when differentiability holds.

More fundamentally, optimal-transport--based distributional objectives exhibit an inherent
sample-wise coupling structure: population level shifts are often driven by a relatively small
subset of influential individuals rather than uniform movement of all samples.
This suggests that practical recourse interventions are naturally budgeted, in the sense
that only a limited number of instances can be meaningfully edited at each step while the
remainder of the empirical distribution should remain unchanged.
Gradient-based solvers, which treat all samples symmetrically, fail to exploit this sparsity
and can therefore produce inefficient or difficult-to-interpret updates.

These challenges motivate the need for a model-agnostic optimization paradigm that (i)
localizes distribution editing under an explicit top-$k$ intervention budget, and (ii) enables
structured search over counterfactual distributions without relying on predictive model
gradients, while preserving the certified DCE objective.

\section{Theoretical Foundations from Distributional Counterfactual Explanations}
\label{sec:appendix_dce_theory}

This section summarizes the key theoretical foundations of DCE that are inherited by DISCOVER.

\subsection{Optimal Transport and Sliced Wasserstein}
\label{sec:appendix_dce_ot}

DCE formulates distributional counterfactual explanations by measuring input proximity and output alignment via optimal transport (OT).
For one-dimensional distributions $\gamma_1,\gamma_2$, the squared 2-Wasserstein distance is defined as
\begin{equation}
\mathcal{W}^{2}(\gamma_1,\gamma_2)
\triangleq
\inf_{\pi\in\Pi(\gamma_1,\gamma_2)}
\int_{\mathbb{R}\times\mathbb{R}}
\|a_1-a_2\|^2\,d\pi(a_1,a_2),
\end{equation}
where $\Pi(\gamma_1,\gamma_2)$ is the set of transport plans with marginals $\gamma_1$ and $\gamma_2$
[\cite{you2025distributional}, Eq.~(1)].

For high-dimensional inputs, DCE uses the sliced Wasserstein distance, which averages 1D Wasserstein distances over random projections:
\begin{equation}
\mathcal{SW}^{2}(\gamma_1,\gamma_2)
\triangleq
\int_{\mathbb{S}^{d-1}}
\mathcal{W}^{2}\!\bigl(\theta_{\sharp}\gamma_1,\theta_{\sharp}\gamma_2\bigr)\,d\sigma(\theta),
\end{equation}
where $\sigma$ is the uniform measure on the unit sphere and $\theta_{\sharp}$ denotes the push-forward under projection
[\cite{you2025distributional}, Eq.~(2)].
This choice preserves a quantile-based interpretation while enabling scalable computation.

DCE casts distributional counterfactual generation as a chance-constrained problem.
Given a factual input distribution $x'$ and a model $b:\mathbb{R}^d\to\mathbb{R}$, let $y'=b(x')$ be the factual output distribution and let $y^\ast$ be a target output distribution.
DCE seeks a counterfactual input distribution $x$ that is close to $x'$ while producing outputs close to $y^\ast$:
\begin{align}
\max_{x,P}\quad & P \\
\text{s.t.}\quad
& P \le \mathbb{P}\!\left[\mathcal{SW}^{2}(x,x') < U_x\right], \\
& P \le \mathbb{P}\!\left[\mathcal{W}^{2}\!\bigl(b(x),y^\ast\bigr) < U_y\right], \\
& P \ge 1-\frac{\alpha}{2},
\end{align}
[\cite{you2025distributional}, Eq.~(3)].
The output-side Wasserstein distance admits an explicit quantile form.
With $y=b(x)$,
\begin{equation}
\mathcal{W}^{2}\!\bigl(y,y^\ast\bigr)
=
\int_0^1
\bigl(F^{-1}_{y}(q)-F^{-1}_{y^\ast}(q)\bigr)^2\,dq,
\end{equation}
which directly compares matched quantiles of the two output distributions
[\cite{you2025distributional}, Eq.~(4)].

DCE also shows that the empirical estimation error rate for $\mathcal{SW}^{2}$ matches that of the 1D $\mathcal{W}^{2}$ estimator under its assumptions.
Formally, if $\mathcal{W}^{2}$ admits sample complexity $\xi(n)$ for 1D empirical estimation, then $\mathcal{SW}^{2}$ admits the same $\xi(n)$ in $\mathbb{R}^d$
[\cite{you2025distributional}, Proposition~3.1].
DISCOVER inherits these OT quantities, since its solver still evaluates candidate distributions using the same $\mathcal{SW}^{2}(x,x')$ and $\mathcal{W}^{2}(b(x),y^\ast)$.

\subsection{Upper Confidence Limits for Certification}
\label{sec:appendix_dce_ucl}

To certify feasibility under finite samples, DCE derives upper confidence limits (UCLs) for both the output-side $\mathcal{W}^{2}(b(x),y^\ast)$ and the input-side $\mathcal{SW}^{2}(x,x')$.
Let $\delta\in(0,\tfrac{1}{2})$ be a trimming constant.
DCE provides a uniform confidence statement (with level $1-\alpha/2$) for the following UCL forms
[\cite{you2025distributional}, Theorem~3.2].

For the output-side Wasserstein term, define
\begin{equation}
\begin{aligned}
D(u)
\triangleq
\max\Bigl\{
& F^{-1}_{y,n}\!\bigl(\bar{q}_{\alpha,n}(u)\bigr)
 - F^{-1}_{y^\ast,n}\!\bigl(\underline{q}_{\alpha,n}(u)\bigr), \\
& F^{-1}_{y^\ast,n}\!\bigl(\bar{q}_{\alpha,n}(u)\bigr)
 - F^{-1}_{y,n}\!\bigl(\underline{q}_{\alpha,n}(u)\bigr)
\Bigr\}.
\end{aligned}
\end{equation}

[\cite{you2025distributional}, Eq.~(7)].
Then, with probability at least $1-\alpha/2$,
\begin{equation}
\mathcal{W}^{2}\!\bigl(b(x),y^\ast\bigr)
\le
\frac{1}{1-2\delta}\int_{\delta}^{1-\delta} D(u)\,du,
\end{equation}
[\cite{you2025distributional}, Eq.~(6)].
DCE denotes the right-hand side as an output-side UCL and checks feasibility by requiring
\begin{equation}
\overline{\mathcal{W}}^{2}
\triangleq
\frac{1}{1-2\delta}\int_{\delta}^{1-\delta} D(u)\,du
\le U_y,
\end{equation}
[\cite{you2025distributional}, Eq.~(10)].

For the input-side sliced Wasserstein term, let $\theta_1,\ldots,\theta_N$ be i.i.d. projection vectors and let $\sigma_N$ be the corresponding empirical measure.
Define
\begin{equation}
\begin{aligned}
D_{\theta,n}(u)
\triangleq
\max\Bigl\{ \;
& F^{-1}_{\theta^\top x,n}\!\bigl(\bar{q}_{\alpha,n}(u)\bigr)
 - F^{-1}_{\theta^\top x',n}\!\bigl(\underline{q}_{\alpha,n}(u)\bigr), \\
& F^{-1}_{\theta^\top x',n}\!\bigl(\bar{q}_{\alpha,n}(u)\bigr)
 - F^{-1}_{\theta^\top x,n}\!\bigl(\underline{q}_{\alpha,n}(u)\bigr)
\; \Bigr\}.
\end{aligned}
\end{equation}
[\cite{you2025distributional}, Eq.~(9)].
Then, with probability at least $1-\alpha/2$,
\begin{equation}
\mathcal{SW}^{2}(x,x')
\le
\frac{1}{1-2\delta}
\int_{\mathbb{S}^{d-1}}
\int_{\delta}^{1-\delta}
D_{\theta,N}(u)\,du\,d\sigma_N(\theta),
\end{equation}
[\cite{you2025distributional}, Eq.~(8)].
DCE denotes the right-hand side as an input-side UCL and checks feasibility by requiring
\begin{equation}
\overline{\mathcal{SW}}^{2}
\triangleq
\frac{1}{1-2\delta}
\int_{\mathbb{S}^{d-1}}
\int_{\delta}^{1-\delta}
D_{\theta,N}(u)\,du\,d\sigma_N(\theta)
\le U_x,
\end{equation}
[\cite{you2025distributional}, Eq.~(11)].

These UCLs operationalize the chance constraints in the DCE formulation while remaining independent of the particular solver.
DISCOVER inherits the same certification step by applying the same UCL computations to candidate counterfactual distributions.

\subsection{Riemannian BCD Objective and Interval Narrowing for $\eta$}
\label{sec:appendix_dce_eta}

DCE solves the chance-constrained formulation by optimizing a weighted objective that balances input proximity and output alignment.
Let $x=\{x_i\}_{i=1}^n$ be an empirical counterfactual input distribution and let $y^\ast=\{y^\ast_j\}_{j=1}^n$ be an empirical target output distribution.
With projection set $\Theta=\{\theta_k\}_{k=1}^N$, DCE defines the empirical OT objectives (we use the standard $1/N$ normalization for the Monte Carlo approximation of $\mathcal{SW}^2$)
\begin{align}
Q_x(x,\mu)
&\triangleq
\frac{1}{N}\sum_{k=1}^{N}\sum_{i=1}^{n}\sum_{j=1}^{n}
\bigl\|\theta_k^\top x_i-\theta_k^\top x'_j\bigr\|^2\,\mu^{(k)}_{ij},
\\
Q_y(x,\nu)
&\triangleq
\sum_{i=1}^{n}\sum_{j=1}^{n}
\bigl\|b(x_i)-y^\ast_j\bigr\|^2\,\nu_{ij},
\end{align}
[\cite{you2025distributional}, Eq.~(12)--(13)].
For a fixed balance parameter $\eta\in[0,1]$, DCE combines them as
\begin{equation}
Q(x,\mu,\nu,\eta)
\triangleq
(1-\eta)\cdot Q_x(x,\mu)+\eta\cdot Q_y(x,\nu),
\end{equation}
[\cite{you2025distributional}, Eq.~(15)].

A key theoretical point in DCE is that, under feasibility, there exists an $\eta^\ast\in[0,1]$ such that minimizing $Q(\cdot,\eta^\ast)$ yields an optimal solution of the original constrained problem
[\cite{you2025distributional}, Theorem~4.1].
This motivates dynamically adapting $\eta$ rather than fixing it.

DCE computes $\eta$ from the UCL gaps and then refines a feasible interval $[l,r]$ across iterations.
Let
\[
a = U_x - \overline{\mathcal{SW}}^{2},\qquad b = U_y - \overline{\mathcal{W}}^{2},
\]
where $\overline{\mathcal{SW}}^{2}$ and $\overline{\mathcal{W}}^{2}$ are the UCLs defined above
[\cite{you2025distributional}, Eq.~(10)--(11)].
DCE specifies the balancing rule
\begin{equation}
\eta =
\begin{cases}
\frac{b}{a+b}, & \text{if $a$ and $b$ are both negative},\\[3pt]
\frac{a}{a+b}, & \text{\parbox[t]{\dimexpr\linewidth-2\tabcolsep\relax}{\raggedright if $a$ and $b$ are both non-negative\\(but not both zero)}},\\[3pt]
0.5, & \text{if $a=b=0$},
\end{cases}
\end{equation}
[\cite{you2025distributional}, Eq.~(25)].
This rule assigns more weight to the side that is either more violated (both negative case) or closer to violation (both non-negative case)
[\cite{you2025distributional}, Appendix~D].

After computing $\eta$, DCE applies Interval Narrowing to update the interval $[l,r]$.
We restate the interval narrowing procedure from DCE for completeness
[\cite{you2025distributional}, Algorithm~2].

\begin{algorithm}[t]
\caption{Interval Narrowing}
\label{alg:interval_narrowing}
\begin{algorithmic}[1]
\REQUIRE$\overline{\mathcal{SW}}^{2}$, $\overline{\mathcal{W}}^{2}$, $U_x,U_y$, $[l,r]$, and $\kappa$ $(0<\kappa<1)$
\ENSURE $\eta$
\STATE $\eta \leftarrow$ Balance the gaps $\overline{\mathcal{SW}}^{2}$ and $\overline{\mathcal{W}}^{2}$
\IF{$\eta > (l + r)/2$}
    \STATE $l \leftarrow l + \kappa(r-l)$
\ELSE
    \STATE $r \leftarrow r - \kappa(r-l)$
\ENDIF
\STATE Save $[l,r]$ and $\kappa$ as the input for the next run
\STATE \textbf{return} $\eta$
\end{algorithmic}
\end{algorithm}

In DCE, the interval $[l,r]$ and $\kappa$ are carried across runs so that $\eta$ is refined progressively rather than changing abruptly
[\cite{you2025distributional}, Algorithm~2 and the discussion around it].
DISCOVER inherits this balancing principle.
Even though DISCOVER replaces gradient-based updates with solver-driven candidate search, it still uses the same two UCL gaps, the same $\eta$ balancing rule, and the same interval narrowing mechanism to maintain the intended trade-off between input proximity and output alignment.

\section{Key Properties of DISCOVER}
\label{sec:appendix_discover_props}

This section provides short proofs for the two key properties stated in the main text.

\subsection{Proof of Proposition 3.1}
By definition,
\[
q_i^{(x)}=\frac{1}{N}\sum_{k=1}^{N}\sum_{j=1}^{n}
\bigl|\theta_k^\top x_i-\theta_k^\top x'_j\bigr|^{2}\,\mu^{(k)}_{ij}.
\]
Summing over $i$ and reordering the summations gives
\[
\sum_{i=1}^n q_i^{(x)}
=
\frac{1}{N}\sum_{k=1}^{N}\sum_{i=1}^{n}\sum_{j=1}^{n}
\bigl|\theta_k^\top x_i-\theta_k^\top x'_j\bigr|^{2}\,\mu^{(k)}_{ij}
= Q_x(x,\mu).
\]
Similarly, using the definition
\[
q_i^{(y)}=\sum_{j=1}^{n}\bigl|b(x_i)-y^\ast_j\bigr|^{2}\,\nu_{ij},
\]
we obtain $\sum_{i=1}^n q_i^{(y)} = Q_y(x,\nu)$.
Finally,
\begin{multline*}
\sum_{i=1}^n q_i
=
(1-\eta)\sum_{i=1}^n q_i^{(x)} + \eta\sum_{i=1}^n q_i^{(y)} \\
=
(1-\eta)Q_x(x,\mu)+\eta Q_y(x,\nu)
=
Q(x,\mu,\nu,\eta).
\end{multline*}
\subsection{Proof of Proposition 3.2}
Let $X^{(0)}=X$ be the no-op proposal included in the candidate set.
By construction,

\begin{align*}
Q(\widehat{X};\eta)
&= \min_{m\in\{0,1,\dots,M\}} Q(X^{(m)};\eta) \\
&\le Q(X^{(0)};\eta) = Q(X;\eta).
\end{align*}
Therefore the propose-and-select step is monotone (non-increasing) in the certified objective for the fixed $\eta$ used during candidate evaluation.

\section{Additional Ablation Results}
\label{sec:appendix_ablation}

Table~\ref{tab:ablation_h_k} reports additional ablations for two solver hyperparameters.
We vary the within-row update size $h$ under the Monte Carlo strategy and the intervention budget $k$ under the Genetic strategy, while keeping the remaining components fixed.

\paragraph{Effect of $h$.}
Increasing $h$ expands the local proposal scope within each selected sample.
Across datasets, larger $h$ tends to reduce $\OTy$ more quickly, but it can increase $\OTx$ and MMD when proposals become less local.
This behavior directly reflects the role of the OT-guided cone sampler: DISCOVER is most effective when proposals remain transport-consistent rather than performing large unconstrained jumps.
The best trade-off often occurs at intermediate values, suggesting that $h$ primarily controls exploration scale within the propose-and-select framework.

\paragraph{Effect of $k$.}
The budget $k$ controls how many samples can be edited per iteration.
On heterogeneous datasets, larger $k$ can reduce $\OTy$ in some settings, but it also increases variance and can worsen $\OTx$.
This supports the core sparsity mechanism of DISCOVER: distributional alignment is achieved through targeted top-$k$ interventions, rather than uniformly shifting the full population.
Overall, these results justify using a small intervention budget to stabilize certified distribution-level search under black-box predictors.

\section{Qualitative Analysis of Distributional Counterfactuals}
\label{sec:appendix_qualitative}

This section provides qualitative visualizations to complement the quantitative evaluation in the main paper.

\subsection{Distributional Shifts across Datasets}
\label{sec:appendix_quantile}

\begin{figure}[t]
    \centering
    \includegraphics[width=0.9\linewidth]{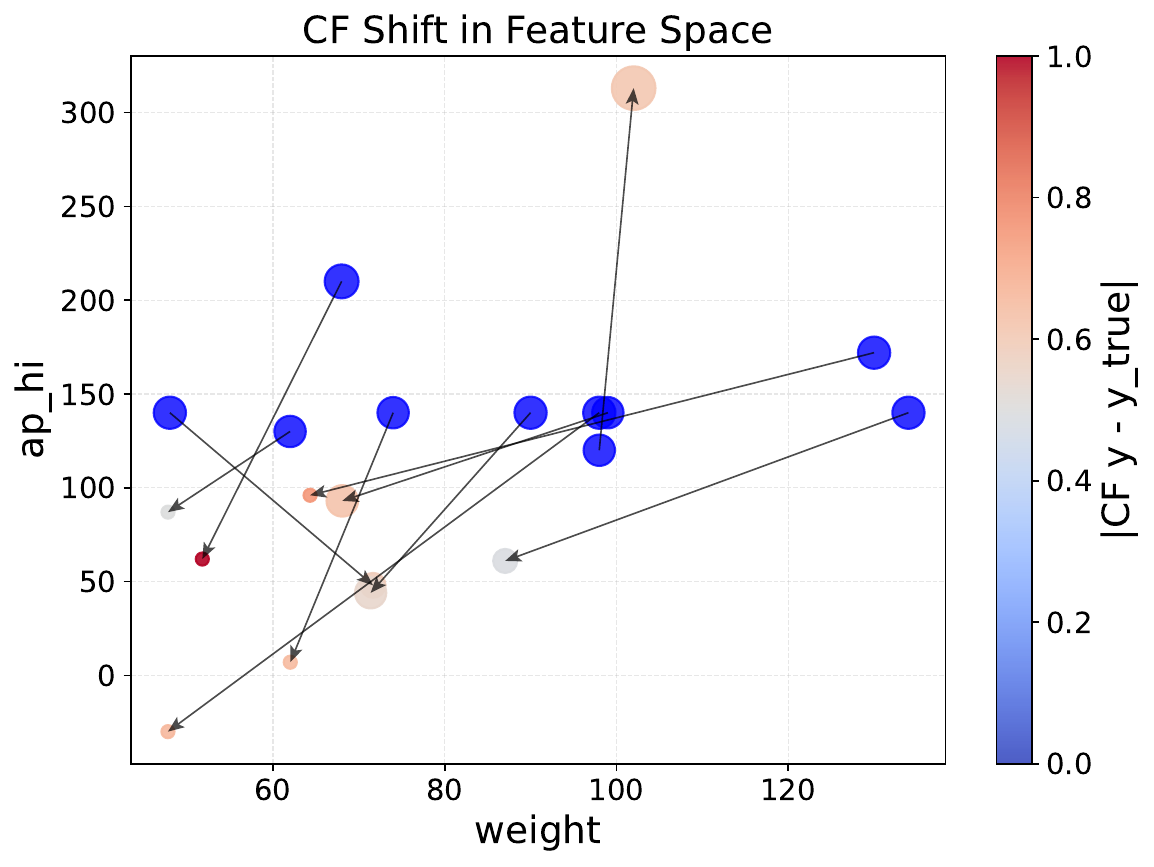}
    \caption{
    Sample-level counterfactual shifts produced by DISCOVER on the Cardio dataset.
    Points denote factual (blue) and counterfactual (red) samples in the weight--systolic blood pressure plane, with arrows indicating the update direction and magnitude.
    Point size encodes diastolic blood pressure (\texttt{ap\_lo}), and color reflects the corresponding reduction in predicted risk.
    }
    \label{fig:cf_shift}
\end{figure}

\begin{figure*}[t]
  \centering
  \includegraphics[width=\textwidth]{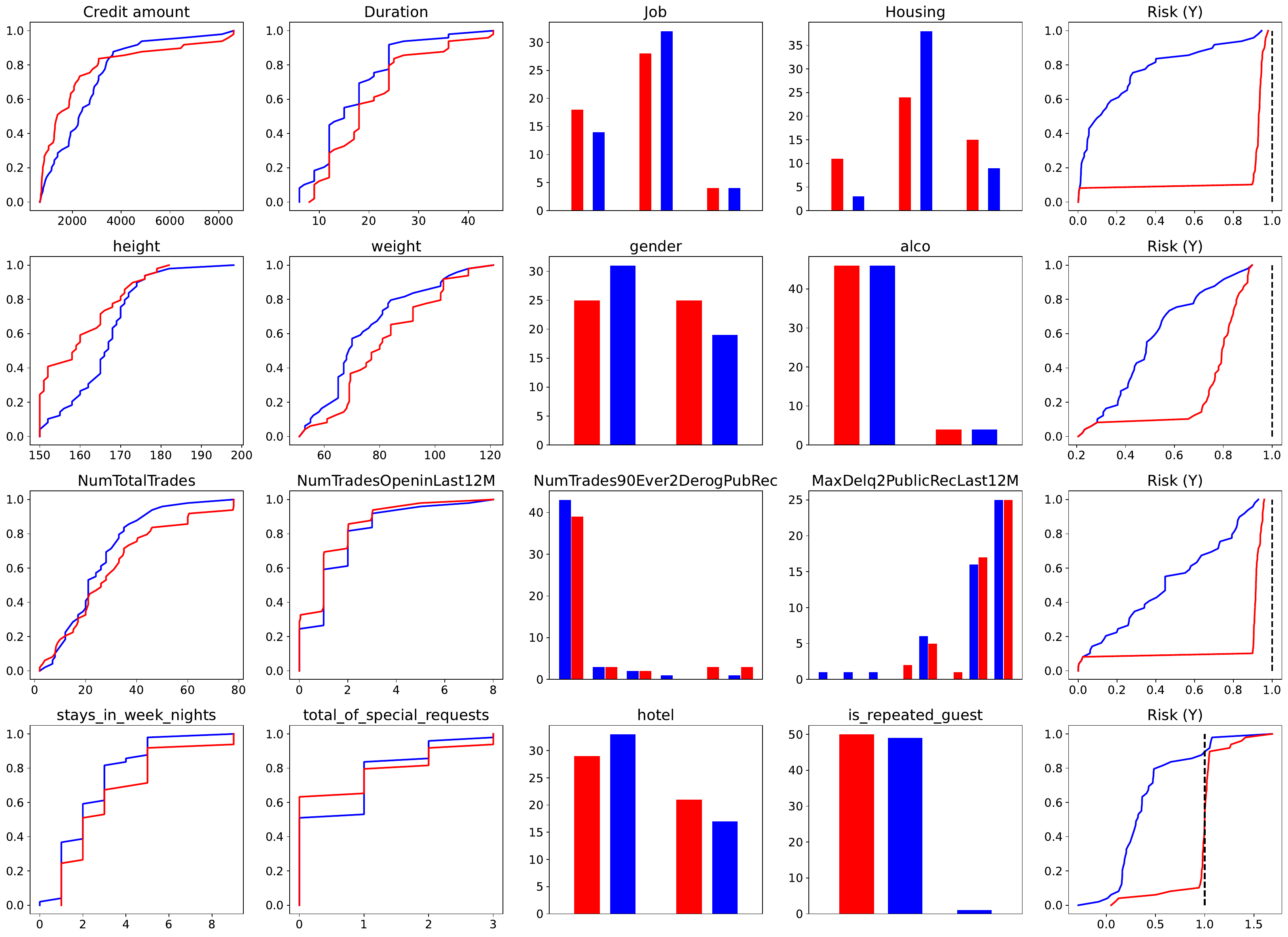}
  \caption{
  Empirical distribution visualizations of factual and counterfactual samples generated by DISCOVER
  on German Credit, Cardio, HELOC, and Hotel Booking.
  For numerical features and model outputs $Y$, curves show empirical CDFs with the horizontal axis representing feature values or risk scores.
  For categorical features, bar plots indicate empirical frequencies.
  Gray curves correspond to factual distributions, dashed black curves denote target output distributions when applicable,
  and colored curves represent DISCOVER counterfactuals.
  }
  \label{fig:qualitative_quantiles}
\end{figure*}

Figure~\ref{fig:qualitative_quantiles} visualizes how DISCOVER reshapes empirical input and output distributions across datasets with mixed feature types.
For numerical features and prediction scores, DISCOVER shifts the empirical CDFs toward the target while largely preserving the overall shape of the factual distributions, which is consistent with low input-side transport distance.
For categorical features, frequency changes are concentrated on a small number of categories, rather than spreading mass broadly across many levels.

Across datasets, the counterfactual outputs move toward the target distribution while the input distributions remain close to the factual reference.
These visual patterns align with the OT-based metrics reported in the main results and provide an interpretable view of distribution-level recourse.

\subsection{Sample-Level Counterfactual Shifts in Feature Space}
\label{sec:qualitative_cf_shift}

\begin{figure}[!t]
  \centering
  \includegraphics[width=0.9\columnwidth]{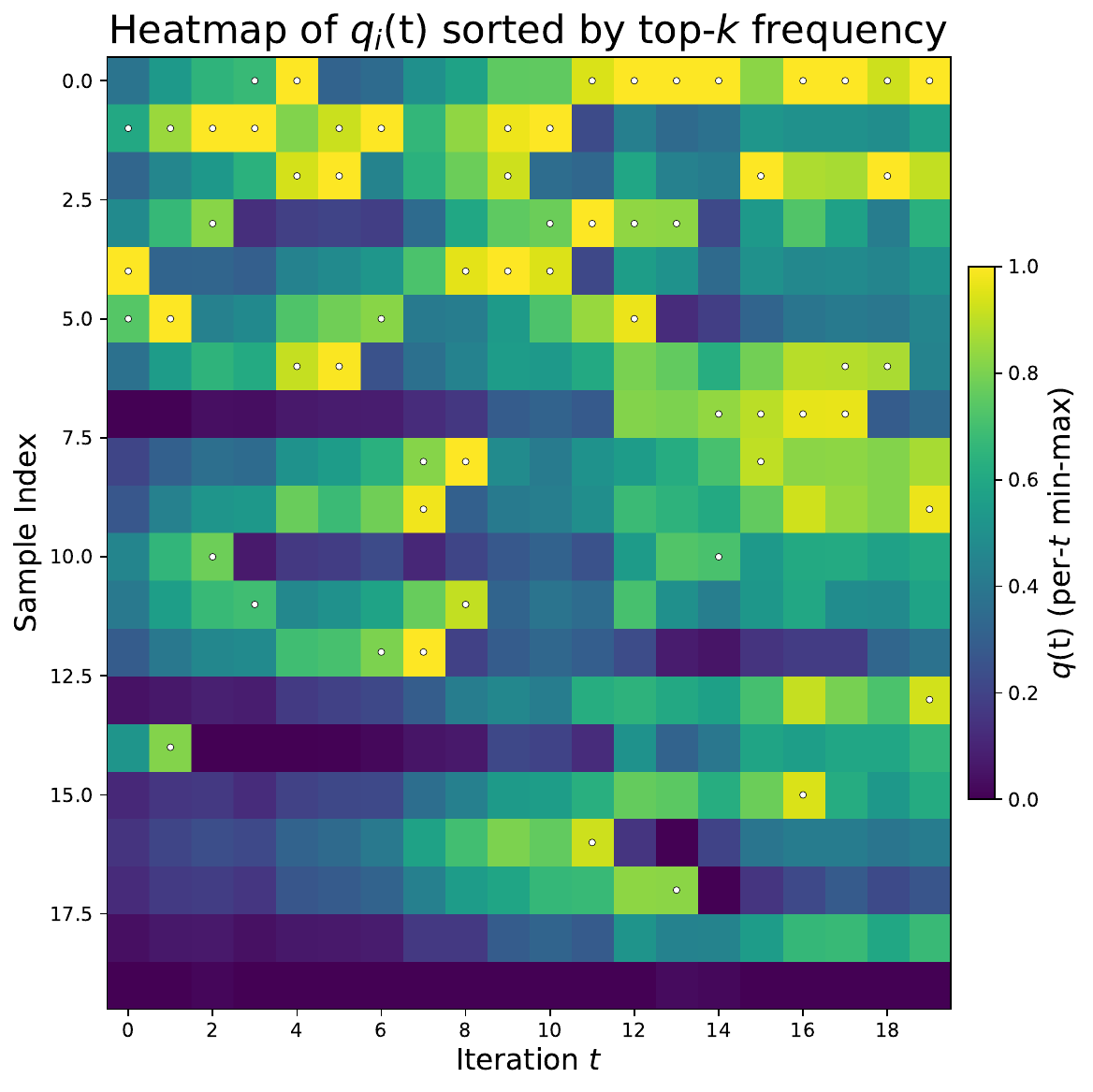}
    \caption{
    Heatmap of $q_i(t)$ with samples sorted by their top-$k$ selection frequency across iterations ($k{=}3$, 20 samples shown).
    The horizontal axis denotes the iteration index $t$.
    The vertical axis corresponds to sample indices after sorting by how frequently each sample appears in the top-$k$ set over all iterations.
    Color encodes $q_i(t)$ using per-iteration min--max normalization, so the colormap emphasizes relative importance within each iteration.
    White circular markers indicate samples selected into the top-$k$ set at iteration $t$.
    }
  \label{fig:qi_heatmap_freq}
\end{figure}

\begin{figure}[!t]
  \centering
  \includegraphics[width=0.9\columnwidth]{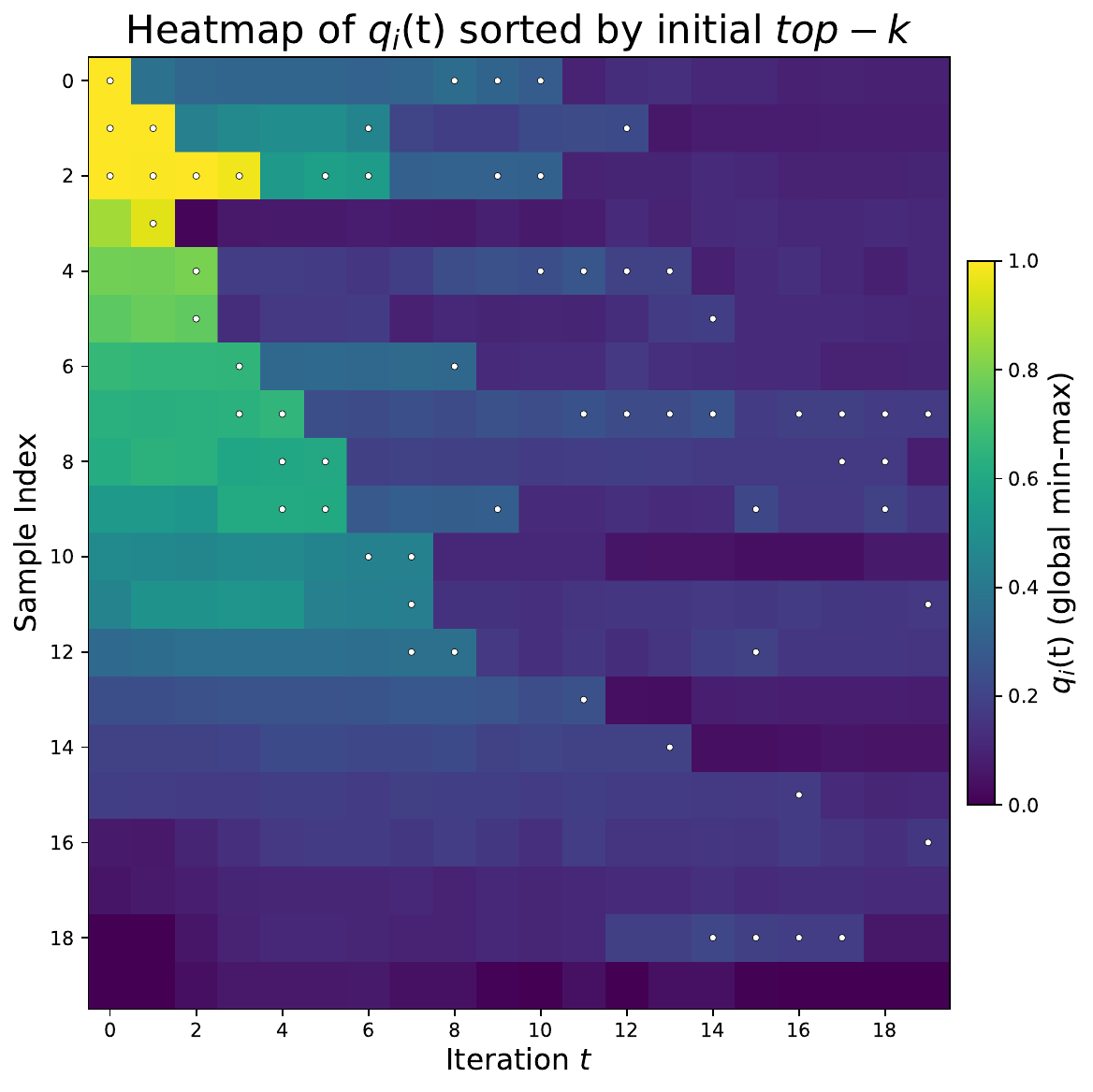}
    \caption{
    Heatmap of $q_i(t)$ with samples sorted by their initial ranking induced by $q_i(0)$ ($k{=}3$, 20 samples shown).
    The horizontal axis denotes the iteration index $t$.
    The vertical axis corresponds to sample indices ordered by the initial ranking, with initial top-$k$ samples appearing at the top.
    Color encodes $q_i(t)$ using a global min--max scale shared across iterations, which highlights changes in absolute magnitude over time.
    White circular markers indicate samples selected into the top-$k$ set at iteration $t$.
    }
  \label{fig:qi_heatmap_init}
\end{figure}

Figure~\ref{fig:cf_shift} illustrates sample-level movements induced by DISCOVER in the input space.
Arrows visualize the change from each factual instance to its counterfactual counterpart, while color encodes the absolute change in model output, $|b(x)-b(x')|$.

Two patterns are consistent with the solver design.
First, most updates remain local, and larger movements are concentrated on a subset of samples, which matches the top-$k$ intervention mechanism.
Second, samples with larger output changes typically exhibit larger input displacements, indicating that distribution-level alignment is achieved through targeted edits rather than uniform shifts of the entire population.

\subsection{Dynamic Reweighting of Sample Contributions}
\label{sec:appendix_heatmaps}

Figure~\ref{fig:qi_heatmap_freq} shows $q_i(t)$ when samples are ordered by long-run top-$k$ selection frequency.
Because the colormap is normalized per iteration, the figure highlights how the solver concentrates attention within each iteration.
A small set of samples is repeatedly selected, which indicates persistent influential contributors under the current transport coupling.
At the same time, non-selected samples still exhibit non-zero scores, since $q_i(t)$ is evaluated for all samples before top-$k$ gating.

Figure~\ref{fig:qi_heatmap_init} shows the same trajectories under an initialization-based ordering and a global color scale.
Early iterations concentrate mass on initially influential samples, but top-$k$ membership changes over time, which indicates that influence is not fixed by initialization.
The global scale also reveals a decrease in the overall magnitude of $q_i(t)$, which is consistent with a reduction in transport discrepancy as optimization progresses.

Overall, these visualizations show that DISCOVER combines global reweighting of sample influence with sparse top-$k$ interventions.
This mechanism concentrates edits on samples that contribute most to the current transport discrepancy while maintaining a stable distribution-level optimization process.

\begin{sidewaystable*}[p!]
\centering
\scriptsize
\setlength{\tabcolsep}{3.0pt}
\renewcommand{\arraystretch}{0.80}
\caption{
Ablation results under Monte Carlo (MC) and Genetic strategies on four datasets.
Panel A varies the step-size parameter $h$ under MC and reports mean $\pm$ 95\% CI over five seeds.
Panel B varies the top-$k$ parameter under the Genetic strategy and reports mean $\pm$ 95\% CI over the three random seeds shared across $k\!\in\!\{3,10,20\}$ (to ensure consistent comparison across $k$ settings).
For each dataset and model, the best value (lowest mean) for each metric within a panel is highlighted in bold.
}
\label{tab:ablation_h_k}
\resizebox{\textwidth}{!}{
\begin{tabular}{cccccccccc}

\toprule

Dataset & $h$ &
\multicolumn{4}{c}{MLP} &
\multicolumn{4}{c}{Random Forest} \\
\cmidrule(lr){3-6}\cmidrule(lr){7-10}
& &
$\OTx$ & $\OTy$ & MMD & AReS Cost &
$\OTx$ & $\OTy$ & MMD & AReS Cost \\
\midrule

\multirow{4}{*}{German Credit}
& $h=1$ & 0.0837 $\pm$ 0.0176 & 0.0196 $\pm$ 0.0307 & \textbf{0.0310 $\pm$ 0.0049} & \textbf{6.59 $\pm$ 0.78}
        & \textbf{0.1102 $\pm$ 0.0281} & 0.1014 $\pm$ 0.0454 & \textbf{0.0353 $\pm$ 0.0038} & \textbf{7.81 $\pm$ 1.31} \\
& $h=2$ & 0.0845 $\pm$ 0.0265 & 0.0058 $\pm$ 0.0072 & 0.0343 $\pm$ 0.0041 & 7.20 $\pm$ 1.13
        & 0.1258 $\pm$ 0.0234 & 0.0680 $\pm$ 0.0207 & 0.0400 $\pm$ 0.0035 & 8.30 $\pm$ 0.58 \\
& $h=4$ & 0.0799 $\pm$ 0.0117 & 0.0019 $\pm$ 0.0011 & 0.0373 $\pm$ 0.0039 & 8.94 $\pm$ 0.61
        & 0.1216 $\pm$ 0.0229 & 0.0571 $\pm$ 0.0217 & 0.0425 $\pm$ 0.0050 & 10.05 $\pm$ 0.81 \\
& $h=8$ & \textbf{0.0778 $\pm$ 0.0183} & \textbf{0.0011 $\pm$ 0.0007} & 0.0458 $\pm$ 0.0088 & 10.54 $\pm$ 1.58
        & 0.1124 $\pm$ 0.0239 & \textbf{0.0372 $\pm$ 0.0109} & 0.0512 $\pm$ 0.0123 & 12.28 $\pm$ 2.51 \\
\midrule

\multirow{4}{*}{COMPAS}
& $h=1$ & \textbf{0.1128 $\pm$ 0.0296} & 0.0313 $\pm$ 0.0083 & \textbf{0.0541 $\pm$ 0.0099} & \textbf{9.95 $\pm$ 2.30}
        & \textbf{0.0559 $\pm$ 0.0176} & 0.0210 $\pm$ 0.0065 & \textbf{0.0247 $\pm$ 0.0078} & 8.42 $\pm$ 3.29 \\
& $h=2$ & 0.1149 $\pm$ 0.0245 & 0.0282 $\pm$ 0.0126 & 0.0564 $\pm$ 0.0053 & 10.24 $\pm$ 2.37
        & 0.0615 $\pm$ 0.0131 & 0.0174 $\pm$ 0.0049 & 0.0284 $\pm$ 0.0063 & 7.75 $\pm$ 1.00 \\
& $h=4$ & 0.1385 $\pm$ 0.0813 & 0.0255 $\pm$ 0.0099 & 0.0553 $\pm$ 0.0122 & 10.63 $\pm$ 2.31
        & 0.0720 $\pm$ 0.0144 & 0.0148 $\pm$ 0.0054 & 0.0318 $\pm$ 0.0085 & 7.17 $\pm$ 0.98 \\
& $h=8$ & 0.1609 $\pm$ 0.0435 & \textbf{0.0169 $\pm$ 0.0083} & 0.0620 $\pm$ 0.0069 & 10.15 $\pm$ 2.56
        & 0.0739 $\pm$ 0.0180 & \textbf{0.0113 $\pm$ 0.0039} & 0.0339 $\pm$ 0.0078 & \textbf{6.96 $\pm$ 1.08} \\
\midrule

\multirow{4}{*}{HELOC}
& $h=1$ & \textbf{0.1108 $\pm$ 0.0220} & 0.0071 $\pm$ 0.0017 & \textbf{0.0340 $\pm$ 0.0053} & 7.25 $\pm$ 0.79
        & \textbf{0.0875 $\pm$ 0.0210} & 0.0291 $\pm$ 0.0171 & \textbf{0.0266 $\pm$ 0.0048} & \textbf{4.48 $\pm$ 0.89} \\
& $h=2$ & 0.1335 $\pm$ 0.0453 & 0.0055 $\pm$ 0.0014 & 0.0393 $\pm$ 0.0038 & 7.37 $\pm$ 0.98
        & 0.0983 $\pm$ 0.0201 & 0.0228 $\pm$ 0.0094 & 0.0321 $\pm$ 0.0058 & 4.96 $\pm$ 1.20 \\
& $h=4$ & 0.1751 $\pm$ 0.0663 & 0.0047 $\pm$ 0.0010 & 0.0405 $\pm$ 0.0041 & 7.55 $\pm$ 1.02
        & 0.1387 $\pm$ 0.0435 & 0.0145 $\pm$ 0.0086 & 0.0337 $\pm$ 0.0065 & 5.36 $\pm$ 0.93 \\
& $h=8$ & 0.1691 $\pm$ 0.0779 & \textbf{0.0035 $\pm$ 0.0013} & 0.0445 $\pm$ 0.0056 & \textbf{7.08 $\pm$ 0.82}
        & 0.1220 $\pm$ 0.0406 & \textbf{0.0129 $\pm$ 0.0093} & 0.0368 $\pm$ 0.0078 & 5.09 $\pm$ 1.06 \\
\midrule

\multirow{4}{*}{Cardio}
& $h=1$ & 0.0643 $\pm$ 0.0425 & 0.0134 $\pm$ 0.0080 & 0.0232 $\pm$ 0.0073 & 7.21 $\pm$ 2.82
        & 0.0369 $\pm$ 0.0140 & 0.0107 $\pm$ 0.0035 & \textbf{0.0192 $\pm$ 0.0039} & \textbf{6.49 $\pm$ 1.00} \\
& $h=2$ & 0.0566 $\pm$ 0.0361 & 0.0109 $\pm$ 0.0089 & 0.0234 $\pm$ 0.0051 & 7.34 $\pm$ 1.55
        & 0.0417 $\pm$ 0.0289 & 0.0075 $\pm$ 0.0027 & 0.0212 $\pm$ 0.0036 & 7.09 $\pm$ 1.57 \\
& $h=4$ & \textbf{0.0534 $\pm$ 0.0285} & \textbf{0.0066 $\pm$ 0.0042} & 0.0265 $\pm$ 0.0069 & 8.48 $\pm$ 1.14
        & \textbf{0.0347 $\pm$ 0.0001} & 0.0071 $\pm$ 0.0039 & 0.0227 $\pm$ 0.0056 & 7.80 $\pm$ 1.67 \\
& $h=8$ & 0.0544 $\pm$ 0.0241 & 0.0067 $\pm$ 0.0023 & \textbf{0.0192 $\pm$ 0.0021} & \textbf{7.13 $\pm$ 1.28}
        & 0.0394 $\pm$ 0.0166 & \textbf{0.0071 $\pm$ 0.0026} & 0.0265 $\pm$ 0.0058 & 8.31 $\pm$ 1.21 \\
\midrule

\addlinespace[3pt] 
\midrule

Dataset & $k$ &
\multicolumn{4}{c}{SVM} &
\multicolumn{4}{c}{XGBoost} \\
\cmidrule(lr){3-6}\cmidrule(lr){7-10}
& &
$\OTx$ & $\OTy$ & MMD & AReS Cost &
$\OTx$ & $\OTy$ & MMD & AReS Cost \\
\midrule

\multirow{3}{*}{German Credit} & $k=3$  & \textbf{0.8351 $\pm$ 0.3131} & \textbf{0.2144 $\pm$ 0.0387} & 0.0602 $\pm$ 0.0201 & 13.96 $\pm$ 1.82 & \textbf{0.0864 $\pm$ 0.0482} & \textbf{0.0033 $\pm$ 0.0031} & \textbf{0.0457 $\pm$ 0.0064} & \textbf{11.35 $\pm$ 0.75} \\
& $k=10$ & 0.8876 $\pm$ 0.5071 & 0.2192 $\pm$ 0.0295 & 0.0237 $\pm$ 0.1022 & \textbf{4.43 $\pm$ 19.08} & 0.1545 $\pm$ 0.0608 & 0.0120 $\pm$ 0.0185 & 0.0467 $\pm$ 0.0015 & 12.81 $\pm$ 3.36 \\
& $k=20$ & 0.8472 $\pm$ 0.4786 & 0.2387 $\pm$ 0.0822 & \textbf{0.0219 $\pm$ 0.0942} & 5.26 $\pm$ 22.63 & 0.2534 $\pm$ 0.3980 & 0.0807 $\pm$ 0.0767 & 0.0537 $\pm$ 0.0128 & 12.59 $\pm$ 1.71 \\
\midrule

\multirow{3}{*}{COMPAS} & $k=3$  & \textbf{0.2660 $\pm$ 0.1678} & \textbf{0.0100 $\pm$ 0.0095} & 0.0866 $\pm$ 0.0021 & \textbf{16.56 $\pm$ 2.81} & \textbf{0.3091 $\pm$ 0.5377} & \textbf{0.0177 $\pm$ 0.0206} & 0.0636 $\pm$ 0.0241 & \textbf{17.37 $\pm$ 9.39} \\
& $k=10$ & 0.4979 $\pm$ 0.1041 & 0.0288 $\pm$ 0.0162 & \textbf{0.0705 $\pm$ 0.0090} & 18.53 $\pm$ 4.48 & 0.3422 $\pm$ 0.4806 & 0.0326 $\pm$ 0.0254 & 0.0795 $\pm$ 0.0318 & 20.35 $\pm$ 6.61 \\
& $k=20$ & 0.4302 $\pm$ 0.3002 & 0.0590 $\pm$ 0.0597 & 0.0899 $\pm$ 0.0084 & 19.03 $\pm$ 7.88 & 0.4080 $\pm$ 0.3655 & 0.0628 $\pm$ 0.0785 & \textbf{0.0592 $\pm$ 0.0284} & 19.67 $\pm$ 3.63 \\
\midrule

\multirow{3}{*}{HELOC} & $k=3$  & 0.1888 $\pm$ 0.1623 & 0.0574 $\pm$ 0.0195 & 0.0620 $\pm$ 0.0051 & \textbf{11.15 $\pm$ 0.88} & \textbf{0.0869 $\pm$ 0.0164} & \textbf{0.0051 $\pm$ 0.0028} & 0.0450 $\pm$ 0.0042 & \textbf{5.72 $\pm$ 0.25} \\
& $k=10$ & \textbf{0.1857 $\pm$ 0.1464} & \textbf{0.0390 $\pm$ 0.0104} & \textbf{0.0554 $\pm$ 0.0065} & 11.77 $\pm$ 1.13 & 0.1185 $\pm$ 0.0284 & 0.0083 $\pm$ 0.0051 & \textbf{0.0395 $\pm$ 0.0030} & 6.12 $\pm$ 0.47 \\
& $k=20$ & 0.3701 $\pm$ 0.3532 & 0.0501 $\pm$ 0.0332 & 0.0700 $\pm$ 0.0067 & 13.46 $\pm$ 3.09 & 0.1003 $\pm$ 0.0331 & 0.0149 $\pm$ 0.0079 & 0.0412 $\pm$ 0.0058 & 6.37 $\pm$ 0.67 \\
\midrule

\multirow{3}{*}{Cardio} & $k=3$  & \textbf{0.0435 $\pm$ 0.0260} & 0.0040 $\pm$ 0.0022 & \textbf{0.0180 $\pm$ 0.0033} & \textbf{10.57 $\pm$ 1.12} & 0.0553 $\pm$ 0.0472 & 0.0033 $\pm$ 0.0025 & 0.0256 $\pm$ 0.0057 & \textbf{11.19 $\pm$ 0.85} \\
& $k=10$ & 0.0558 $\pm$ 0.0304 & \textbf{0.0035 $\pm$ 0.0021} & 0.0209 $\pm$ 0.0062 & 11.23 $\pm$ 1.29 & \textbf{0.0431 $\pm$ 0.0329} & \textbf{0.0027 $\pm$ 0.0015} & \textbf{0.0219 $\pm$ 0.0044} & 12.47 $\pm$ 1.55 \\
& $k=20$ & 0.0619 $\pm$ 0.0480 & 0.0046 $\pm$ 0.0031 & 0.0268 $\pm$ 0.0072 & 13.20 $\pm$ 1.48 & 0.0502 $\pm$ 0.0388 & 0.0043 $\pm$ 0.0038 & 0.0253 $\pm$ 0.0058 & 13.29 $\pm$ 1.82 \\

\bottomrule
\end{tabular}
}
\end{sidewaystable*}

\subsection{Computational Complexity of DISCOVER}
\label{sec:complexity}

\paragraph{Notation.}
Let $n$ be the number of samples in the empirical distribution, $d$ the number of input features,
$N$ the number of random projection directions used to approximate the sliced Wasserstein distance,
$T$ the number of outer iterations, $M$ the number of candidate proposals per iteration,
$k$ the top-$k$ intervention budget (number of editable rows per iteration),
and $h$ the number of edited features per selected row (a within-row update budget).
For categorical feature $p$, let $|V_p|$ be its number of discrete values and let $r$ be the embedding dimension used in the
categorical proposal step. Let $C_b$ denote the cost of one forward call to the predictor $b(\cdot)$ on a single sample
(model-query cost; for vectorized inference $C_b$ can be interpreted as amortized per-sample cost).

\paragraph{One-time precomputation.}
Because the factual distribution $X'$ and target outputs $Y^\ast$ are fixed, we can precompute:
(i) for each projection $\theta_k$ ($k=1,\dots,N$), the projected factual scalars $\{\theta_k^\top x'_j\}_{j=1}^n$
and their sorted order; and
(ii) the sorted target outputs $\{y^\ast_j\}_{j=1}^n$.
This costs
\[
\mathcal{O}\!\left(N(nd + n\log n) + n\log n\right)
\]
time and stores $\mathcal{O}(Nn)$ projected scalars (plus indices).

\paragraph{Cost of evaluating the certified objective on one distribution.}
Evaluating the certified objective $Q(X;\eta)=(1-\eta)Q_x(X)+\eta Q_y(X)$ on a candidate empirical distribution $X$
requires:

\begin{itemize}
\item \textbf{Input-side $SW_2$:}
for each projection $\theta_k$, compute $\{\theta_k^\top x_i\}_{i=1}^n$ (\,$\mathcal{O}(nd)$\,),
sort the projected values (\,$\mathcal{O}(n\log n)$\,), and compute the 1D OT cost and UCL terms
via rank-matching (\,$\mathcal{O}(n)$\,).
Thus, for $N$ projections,
\[
\mathcal{O}\!\left(N(nd + n\log n)\right).
\]

\item \textbf{Output-side $W_2$:}
evaluate the predictor on all samples $y_i=b(x_i)$ (\,$\mathcal{O}(nC_b)$\,),
sort $\{y_i\}_{i=1}^n$ (\,$\mathcal{O}(n\log n)$\,),
and compute the 1D OT cost and UCL terms (\,$\mathcal{O}(n)$\,).
Overall,
\[
\mathcal{O}\!\left(nC_b + n\log n\right).
\]
\end{itemize}

Therefore, a full evaluation of $Q(X;\eta)$ (including the UCL-based feasibility checks used to update $\eta$) costs
\begin{equation}
\label{eq:cost_Q_full}
\mathcal{O}\!\left(N(nd + n\log n) + nC_b + n\log n\right).
\end{equation}

\paragraph{Per-iteration cost of DISCOVER (worst-case, recomputing from scratch).}
At outer iteration $t$, DISCOVER performs:

\begin{enumerate}
\item \textbf{Evaluate $Q(X_t;\eta_t)$ and update $\eta_t$:} one call to~\eqref{eq:cost_Q_full}.
\item \textbf{Compute per-row impact scores and select top-$k$:}
given the rank-matching structure produced while computing $SW_2$ and $W_2$,
the per-row contributions can be accumulated in $\mathcal{O}(Nn+n)$ time (plus a top-$k$ selection
that is $\mathcal{O}(n)$ expected via selection, or $\mathcal{O}(n\log n)$ if fully sorting).
This does not change the leading-order term of~\eqref{eq:cost_Q_full}.
\item \textbf{Compute the input-side guidance field $g$:}
if we materialize a full gradient-like field $g\in\mathbb{R}^{n\times d}$ from the sliced OT matches,
the accumulation across $N$ projections costs $\mathcal{O}(Nnd)$.
(If guidance is only needed on the editable set $I$ of size $k$, this reduces to $\mathcal{O}(Nkd)$; see below.)
\item \textbf{Generate $M$ candidates (proposal step):}
each proposal edits only $k$ rows and $h$ features per row.
For numerical edits, this costs $\mathcal{O}(Mkh)$.
For categorical edits, decoding may require scanning all categories in edited features, yielding
$\mathcal{O}\!\left(Mk \sum_{p\in\mathcal{C}_{\text{edit}}} |V_p|\,r\right)$ in the worst case,
where $\mathcal{C}_{\text{edit}}$ denotes the set of edited categorical features.
\item \textbf{Select the best candidate via certified objective:}
naively, DISCOVER evaluates $Q(\cdot;\eta_t)$ for each candidate $X_t^{(m)}$ ($m=1,\dots,M$),
which costs $M$ additional calls to~\eqref{eq:cost_Q_full}.
\end{enumerate}

Putting the dominant terms together, the worst-case per-iteration time complexity is

\begin{equation}
\begin{split}
\mathcal{O}\!\left( (M+1)\Big(N(nd + n\log n) + nC_b + n\log n\Big) \right. \\
\left. + Nnd + Mkh \right),
\end{split}
\end{equation}
where the $Nnd$ term corresponds to computing a dense guidance field $g$.
In practice, the proposal-generation cost $Mkh$ is typically dominated by candidate scoring
unless $M$ is extremely small.

Hence the total worst-case runtime over $T$ iterations is

\begin{equation*}
\begin{split}
\mathcal{O}\!\left( T(M+1) \Big( N(nd + n\log n) \right. & \left. + nC_b + n\log n \Big) \right) \\
& \quad\text{(dominant term).}
\end{split}
\end{equation*}

\begin{table*}[t]
\centering
\caption{Runtime analysis of DISCOVER on COMPAS with a LightGBM classifier. Runtime is reported as mean $\pm$ standard deviation (seconds) over five runs.}
\label{tab:runtime}
\setlength{\tabcolsep}{4pt}

\begin{tabular}{lccccc|lccccc}
\toprule

\multicolumn{6}{c|}{Sample Size ($n$)}
&
\multicolumn{6}{c}{Top-$k$}
\\

\midrule

$n$
& 25 & 50 & 100 & 150 & 200
&
$k$
& 1 & 2 & 3 & 5 & 10
\\

Runtime
& 34.8
& 61.5
& 146.1
& 322.3
& 499.3
&
Runtime
& 35.3
& 35.8
& 36.7
& 37.9
& 41.2
\\

Std
& 4.1
& 2.4
& 9.7
& 25.5
& 23.9
&
Std
& 2.3
& 2.4
& 2.5
& 2.5
& 2.9
\\

\midrule

\multicolumn{6}{c|}{Cone Parameter ($h$)}
&
\multicolumn{6}{c}{Trials ($T$)}
\\

\midrule

$h$
& 1 & 2 & 3 & 5 & 8
&
$T$
& 5 & 10 & 20 & 40 & 80
\\

Runtime
& 35.4
& 35.4
& 35.6
& 35.7
& 36.1
&
Runtime
& 25.8
& 28.8
& 35.3
& 48.4
& 74.4
\\

Std
& 2.3
& 2.3
& 2.3
& 2.3
& 2.4
&
Std
& 1.2
& 1.6
& 2.4
& 3.6
& 6.5
\\

\bottomrule
\end{tabular}

\end{table*}

\paragraph{Sparsity-aware amortized cost (recommended implementation).}
The top-$k$ intervention budget implies that each candidate differs from $X_t$ in only $k$ rows.
This permits two practical optimizations that reduce constant factors and, with additional data structures,
can also reduce asymptotic costs:

\begin{itemize}
\item \textbf{Sparse guidance:} compute guidance only on the editable set $I$,
reducing the guidance construction from $\mathcal{O}(Nnd)$ to $\mathcal{O}(Nkd)$.

\item \textbf{Incremental model queries:} because only $k$ rows change, one can cache $y_i=b(x_i)$ for the current iterate
and re-evaluate $b(\cdot)$ only on edited rows for each candidate, reducing model calls from $\mathcal{O}(nC_b)$ to
$\mathcal{O}(kC_b)$ per candidate. The output-side sorting step remains $\mathcal{O}(n\log n)$ if recomputed from scratch,
but can be updated in $\mathcal{O}(k\log n)$ if maintaining an order-statistics structure.

\item \textbf{Incremental projections:} similarly, projected values $\theta_k^\top x_i$ can be cached and updated only for edited rows,
reducing projection computation from $\mathcal{O}(Nnd)$ to $\mathcal{O}(Nkd)$ per candidate. As above, sorting can be
recomputed from scratch (\,$\mathcal{O}(Nn\log n)$\,) or incrementally updated (\,$\mathcal{O}(Nk\log n)$\,) with
balanced-tree / order-statistics maintenance per projection.
\end{itemize}

Under caching and incremental updates with recomputed sorting (a conservative middle ground),
the per-candidate evaluation cost becomes
\[
\mathcal{O}\!\left(N(kd + n\log n) + kC_b + n\log n\right),
\]
while the dense-from-scratch baseline remains~\eqref{eq:cost_Q_full}.
If incremental sorting is used, the $n\log n$ terms can be replaced by $k\log n$ in both the input and output sides.

Storing the current distribution $X\in\mathbb{R}^{n\times d}$ requires $\mathcal{O}(nd)$ memory.
Caching projected scalars for $N$ projections requires $\mathcal{O}(Nn)$ memory, and storing the predictor outputs
requires $\mathcal{O}(n)$. Per-row scores require $\mathcal{O}(n)$.
A dense guidance field $g\in\mathbb{R}^{n\times d}$ costs $\mathcal{O}(nd)$ memory, but can be avoided by computing guidance only
for the editable set, reducing storage to $\mathcal{O}(kd)$.

The DISCOVER algorithm in the main paper is written as if all $M$ candidates are materialized; that would require $\mathcal{O}(Mnd)$ memory.
However, a streaming implementation that generates and scores candidates one-by-one only needs to keep the best-so-far candidate,
reducing memory to $\mathcal{O}(nd + Nn)$ (plus caches and small proposal buffers).

\begin{table*}[t]
\centering
\caption{Performance comparison of model agnostic counterfactual explanation methods on COMPAS using SVM and MLP.}
\label{tab:compas_model_agnostic_cf_comparison}
\setlength{\tabcolsep}{4pt}
\begin{tabular}{lcccccccc}
\toprule
\multirow{2}{*}{Method}
& \multicolumn{4}{c}{SVM}
& \multicolumn{4}{c}{MLP} \\
\cmidrule(lr){2-5} \cmidrule(lr){6-9}
& $\mathrm{OT}(x)$
& $\mathrm{OT}(y)$
& MMD
& AReS Cost
& $\mathrm{OT}(x)$
& $\mathrm{OT}(y)$
& MMD
& AReS Cost \\
\midrule
AReS
& $0.078$
& $0.503$
& $0.075$
& $3.391$
& $\mathbf{0.014}$
& $0.625$
& $\mathbf{0.016}$
& $1.871$ \\

GLOBE
& $37.127$
& $\mathbf{0.018}$
& $0.226$
& $\mathbf{1.658}$
& $3.124$
& $\mathbf{0.000}$
& $0.238$
& $\mathbf{0.464}$ \\

DiCE
& $0.085$
& $0.244$
& $0.051$
& $3.933$
& $0.104$
& $0.199$
& $0.070$
& $3.973$ \\

DCE
& $0.100$
& $0.167$
& $0.082$
& $3.689$
& $0.064$
& $0.160$
& $0.070$
& $3.354$ \\

NICE
& $0.117$
& $0.261$
& $0.087$
& $3.122$
& $0.156$
& $0.268$
& $0.102$
& $2.917$ \\

MAPOCAM
& $0.272$
& $0.172$
& $0.148$
& $4.059$
& $0.281$
& $0.126$
& $0.161$
& $3.858$ \\

DISCOVER
& $\mathbf{0.076}$
& $0.235$
& $\mathbf{0.048}$
& $3.175$
& $0.147$
& $0.099$
& $0.092$
& $4.427$ \\
\bottomrule
\end{tabular}%
\end{table*}

\section{Reproducibility and Configuration Details}

To support reproducibility, experiments in this work are driven by explicit configuration files and executable scripts. All experiments can be reproduced. In particular, baseline comparison experiments are executed through provided notebooks, while ablation studies are conducted using Python scripts with structured JSON-based configuration files. Baseline methods are evaluated using their original implementations, following the default or recommended parameter settings provided by the respective authors. Consequently, baseline hyper-parameters are not globally fixed across datasets, which is consistent with common practice in counterfactual explanation benchmarks.

For DISCOVER, parameter choices are structured to isolate the effect of individual components. In ablations of the OT-guided cone sampling mechanism, both the intervention budget and the within-sample update budget are fixed to $k = 1$ and $h = 3$, respectively. This ensures that observed differences are attributable solely to the presence or absence of cone guidance. Cone sampling is evaluated under four variants: full guidance, continuous-only, categorical-only, and no guidance, using a shared candidate generation budget and the Monte Carlo proposal strategy on all five datasets.

In experiments analyzing sparse intervention budgets, the cone sampling configuration is held fixed and applied uniformly to both continuous and categorical features. We then study the effects of top-$k$ and $h$ through separate controlled sweeps. Specifically, in the $h$-ablation, the top-$k$ intervention budget is kept fixed while varying the number of features updated within each selected sample over $h \in \{1, 2, 4, 8\}$. Conversely, in the top-$k$ ablation, the within-sample update budget $h$ is fixed while varying the number of edited samples per iteration over top-$k \in \{3, 10, 20\}$. These ranges are chosen to cover both conservative and more aggressive intervention regimes while maintaining sparse but non-trivial distributional updates. Both Monte Carlo and Genetic optimizers are evaluated under these budgeted configurations on four datasets.

All quantitative results are averaged over multiple random seeds. Cone-sampling ablations use three independent runs per setting, while experiments involving top-$k$ and $h$ variations use five independent runs per setting to ensure stability of the reported distribution-level OT metrics.
Each run starts from $n=50$ factual samples and returns one empirical counterfactual distribution if the final DCE-style feasibility check is satisfied. For binary classifiers, target outputs are constructed using the decision threshold of the corresponding trained classifier, set to $\tau_{\mathrm{cls}}=0.5$ in our experiments.

\section{Runtime and Scalability Analysis}
 
We further report an empirical runtime analysis of DISCOVER on COMPAS with a LightGBM classifier. Runtime is reported as mean $\pm$ sample standard deviation over five repeated executions. The results show that runtime mainly increases with the sample size and the number of trials. In contrast, increasing the top-$k$ budget or the cone parameter $h$ only introduces a modest runtime increase under the default setting. This provides an empirical profile of the computational cost of DISCOVER under different search budgets and input sizes.

\section{Additional Model Agnostic Baselines}

We additionally compare DISCOVER with NICE~\cite{brughmans2021nice} and MAPOCAM~\cite{raimundo2024mining}, two model agnostic counterfactual explanation methods based on non differentiable search. Table~\ref{tab:compas_model_agnostic_cf_comparison} reports the results on COMPAS with SVM and MLP classifiers. On SVM, DISCOVER achieves the lowest $\mathrm{OT}(x)$ and MMD among all methods. On MLP, DISCOVER achieves the lowest $\mathrm{OT}(y)$ and improves over NICE and MAPOCAM on $\mathrm{OT}(x)$, $\mathrm{OT}(y)$, and MMD. Overall, these results suggest that DISCOVER provides modest improvements over existing model agnostic search baselines in terms of distributional counterfactual quality.

\end{document}